\documentclass[sigconf,10pt]{acmart}

\usepackage{xcolor}
\usepackage{xurl}
\usepackage{hyperref}
\usepackage{booktabs}
\usepackage[ruled]{algorithm2e}
\usepackage{algpseudocode}
\usepackage{enumerate}
\usepackage[english]{babel}
\usepackage{blindtext}
\usepackage{subcaption}
\usepackage{bm}
\usepackage{amsmath}

\usepackage{amssymb}
\usepackage{xspace}
\usepackage{multirow}
\usepackage{threeparttable}
\usepackage{float}
\usepackage{flafter}
\usepackage{comment}
\usepackage[skip=12pt]{caption}
\usepackage{subcaption}
\usepackage{graphicx}
\usepackage{ulem}
\usepackage{verbatim}
\usepackage{array}
\usepackage{enumitem}
\usepackage{booktabs}
\usepackage{multirow}
\usepackage{colortbl}
\usepackage{lipsum}
\usepackage{marvosym}

\def\fig{Fig.\xspace}
\def\eqn{Eq.\xspace}

\def\tab{Tab.\xspace}

\newcommand{\head}[1]{{\noindent \textbf{#1:}}}
\newcommand{\shead}[1]{{\noindent \textbf{#1}}}

\graphicspath{{./figures/}}
\graphicspath{{./pdf/}}
\DeclareGraphicsExtensions{.pdf,.jpeg,.png}

\def\sysname{\textit{MindGuard}\xspace}

\ifodd 1
\newcommand{\com}[1]{\textbf{\color{red}(COMMENT: #1)}} 
\newcommand{\todo}[1]{\textbf{{\color{orange}(TODO: #1)}}}
\newcommand{\unused}[1]{{\color{gray}#1}}
\newcommand{\sheng}[1]{\textbf{\color{olive}(Sheng: #1)}} 
\else

\newcommand{\com}[1]{}
\newcommand{\todo}[1]{}
\newcommand{\unused}[1]{}
\newcommand{\sheng}[1]{}

\fi

\renewcommand\footnotetextcopyrightpermission[1]{} 
\setcopyright{none}

\begin{document}

\title[\sysname]{\sysname: Towards Accessible and Sitgma-free Mental Health First Aid via Edge LLM
}

\author{Sijie Ji$^{\ast \dagger \ddag}$, Xinzhe Zheng$^{\ast}$, Jiawei Sun, Renqi Chen,
Wei Gao$^{\dagger}$, Mani Srivastava$^{\ddag}$}

\affiliation{%
  \institution{$^\dagger$ California Institute of Technology, $\ddag$ University of California, Los Angeles, 
  }
 \country{}
}

\thanks{$^{\ast}$ These authors contribute equally.}

\renewcommand{\shortauthors}{\sysname}

\begin{abstract}
Mental health disorders are among the most prevalent diseases worldwide, affecting nearly one in four people.
Despite their widespread impact, the intervention rate remains below 25\%, largely due to the significant cooperation required from patients for both diagnosis and intervention.
The core issue behind this low treatment rate is stigma, which discourages over half of those affected from seeking help.
This paper presents \sysname, an accessible, stigma-free, and professional mobile mental healthcare system designed to provide mental health first aid.
The heart of \sysname is an innovative edge LLM, equipped with professional mental health knowledge, that seamlessly integrates objective mobile sensor data with subjective Ecological Momentary Assessment records to deliver personalized screening and intervention conversations.
We conduct a broad evaluation of \sysname using open datasets spanning four years and real-world deployment across various mobile devices involving 20 subjects for two weeks.
Remarkably, \sysname achieves results comparable to GPT-4 and outperforms its counterpart with more than 10 times the model size.
We believe that \sysname paves the way for mobile LLM applications, potentially revolutionizing mental healthcare practices by substituting self-reporting and intervention conversations with passive, integrated monitoring within daily life, thus ensuring accessible and stigma-free mental health support.

\end{abstract}

\maketitle

\section{Introduction}

\begin{figure}[t]
  
  \includegraphics[width=0.45\textwidth]{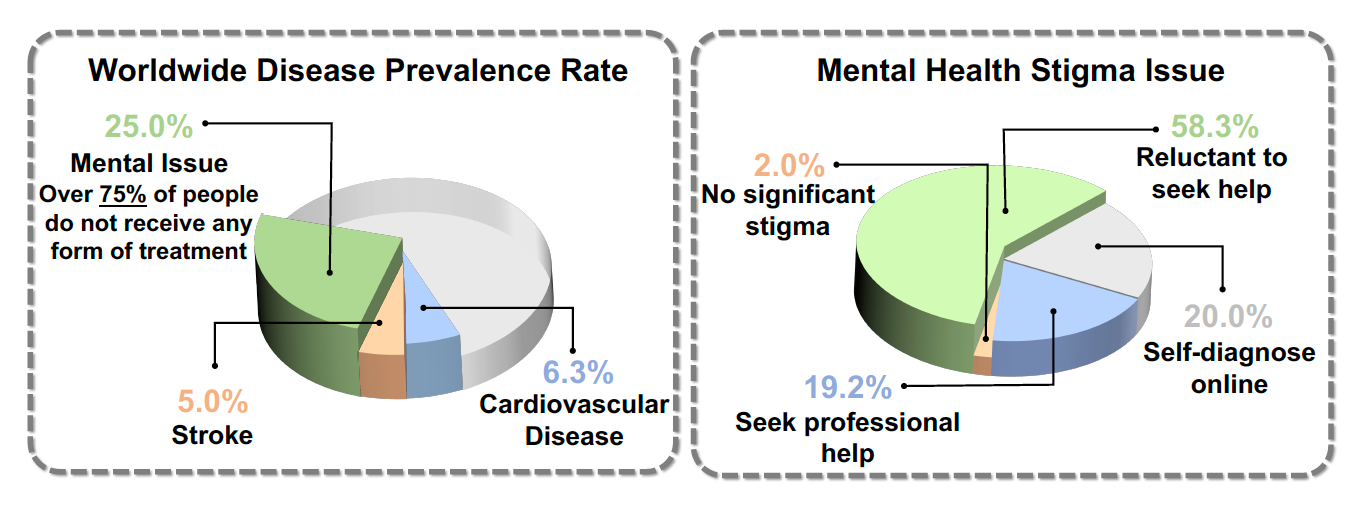}
  \vspace{-2mm}
  \caption{Statistical information on mental health disorders' prevalence and treatment rate.}\label{fig:first-graph}
  
\end{figure}

In fact, mental health disorders (e.g., anxiety, depression and etc.) have the highest prevalence rates compared to many other major health conditions, such as cardiovascular diseases and stroke.
According to the World Health Organization (WHO), approximately 25\% of people globally experience a mental health issue~\cite{world2019mental}, while the prevalence rate for cardiovascular diseases is 6.2\%~\cite{heart}, and for stroke, it is 5\%~\cite{stroke2022world} (\fig\ref{fig:first-graph}).
Moreover, the COVID-19 pandemic and other global crises such as wars and economic downturns have exacerbated this issue, leading to a 25.6\% increase in anxiety and a 27.6\% rise in depression since 2020~\cite{mahmud2023global}. 
Even more concerning is that people with mental illnesses have alarmingly low rates of access to medical treatment~\cite{10665-254610}.
In the United States, only about 36.9\% of adults with a mental health disorder received treatment in the past year, with the rates for teenagers even lower at around 30\%~\cite{terlizzi2020mental}. 
The situation is even more dire in low- and middle-income countries, where WHO estimates that over 75\% of people with severe mental health disorders do not receive any form of treatment~\cite{kohn2004treatment,treatmentrates}.
Only 16.5\% of individuals with depression worldwide seek help and receive minimally adequate treatment~\cite{hooten2016chronic,thornicroft2017undertreatment}.
The widespread nature of mental health issues and their severe inaccessibility to care, combined with their chronic tendencies and long-term impacts, underscores the urgent need to address this significant public health challenge~\cite{vos2020global}.

The primary cause of low treatment rates and the biggest barrier to delivering mental health care is attributed to stigma, which refers to the negative attitudes and discrimination directed towards individuals with their mental illness~\cite{wahl1999mental, who2022stigma}. 
Unlike other diseases, both the diagnosis and intervention of mental illness require substantial user cooperation through conversation.
However, this process usually encounters a high nonresponse rate due to stigma.

As revealed by a comprehensive survey involving over 90,000 participants worldwide, an overwhelming 98\% of survey participants acknowledged the significant stigmatization faced by individuals with mental health disorders and 60\% of them are reluctant to seek professional help due to stigma~\cite{corrigan2000mental}.
In contrast, it is interesting to note that about 20\% of people with mental illnesses report having taken the initiative to self-diagnose and seek help on the Internet because it makes them feel safer and stigma-free~\cite{gould2002seeking,kruzan2022wanted}. 
Unfortunately, existing online tools typically offer only closed-ended questions and construct conversations with limited assessment capabilities, lacking personalized interaction, continuous monitoring, and intervention. 
Existing research efforts primarily focus on delivering better intervention conversations~\cite{nie2024llm,fitzpatrick2017delivering} and engaging users more effectively through mobile devices~\cite{schroeder2018pocket,wang2015using}, which typically occur after individuals have actively sought diagnosis and help—a process that often has a low participation rate. However, there is a gap in utilizing behavioral data to reflect mental health statuses and leveraging mobile devices to facilitate daily conversations for mental health first aid (MHFA). This approach could provide continuous monitoring and early intervention, addressing the critical low treatment rate of mental health caused by stigma.
To fill this gap, this paper presents \sysname, an LLM-powered mobile mental healthcare system that provides accessible MHFA to a huge portion of people in a stigma-free fashion. 
\sysname offers a holistic full-stop solution that facilitates the identification of mental health conditions, enables continuous monitoring, and reframes thoughts and situations through personalized reflective listening and professionally assisted intervention.

At the core of \sysname is the utilization of the causal relationship between user behavior and mental status.
By combining the objective user behavior data collected from the rich sensors on mobile devices with the advanced conversational capabilities of large language models (LLMs), \sysname provides open-ended questions to comprehensively assess the user’s mental state and further deliver personalized conversations and interventions.
Achieving such a system is non-trivial and faces three major challenges. 

First, due to the autoregressive nature of LLMs, \sysname must address the challenge of hallucinations to ensure it consistently delivers accurate and reliable mental health knowledge.
The issue of hallucinations—where the model produces plausible-sounding but incorrect or inaccurate information—stems from its reliance on statistical pattern matching for content generation, limitations in domain-specific training data, the absence of fact-checking capabilities, and a tendency to overgeneralize in complex or rare situations.
\sysname addresses this challenge through a comprehensive approach, including the construction of a high-quality mental health dataset, fine-tuning the LLM using continuous pertaining (PT) techniques~\cite{gururangan2020don}, optimizing the model to follow desired answers through supervised fine-tuning (SFT)~\cite{zhang2023instruction}, and incorporating post-processing with human-in-the-loop review.
This full-chain strategy significantly enhances the LLM's performance in the mental health domain, whereas relying solely on prompt engineering is insufficient to fully resolve the hallucination issue.
The second challenge is how to interpret sensor data in the context of mental health for the LLM model, particularly since the sensor data is in digital form, often noisy, incomplete, and subject to varying behavioral patterns associated with different mental health issues.
\sysname addresses this by proposing a self-refinement and self-feedback mechanism to format the sensor data, allowing the LLM to iteratively update and optimize the data format based on its own feedback, thereby enhancing its ability to accurately interpret and utilize behavioral insights.
Additionally, \sysname introduces counterfactual augmentation learning, which generates misleading information to challenge and improve the LLM's robustness. 
The third challenge is to enable the system to run on mobile devices with acceptable latency to mitigate the risk of sensitive privacy data breaches.
To reduce the size of \sysname without compromising its capabilities, we employ a teacher-student knowledge distillation paradigm.
This approach transfers advanced reasoning skills from state-of-the-art LLMs, such as GPT-4o~\cite{openai2024gpt4o}, to \sysname.
Following this, we utilize the MLC-LLM~\cite{mlc-llm} deployment framework to implement a quantized q4f16 version of \sysname, making it compatible with a wide range of mobile devices.
We summarize the contributions of this paper as follows:
\begin{itemize}[leftmargin=*]
    \item To the best of our knowledge, \sysname is the first system to integrate objective mobile sensor data with subjective LLM-powered conversational data to create a comprehensive mental healthcare system. Unlike existing research that primarily focuses on improving intervention performance, \sysname prioritizes providing accessible and stigma-free MHFA. It offers a holistic solution that encompasses early diagnosis, continuous monitoring, and personalized intervention.
    \item \sysname introduces a suite of fine-tuning methods and strategies designed to address the critical challenge of hallucinations in LLMs—a frequent issue where models produce plausible yet incorrect or misleading information. Experimental results demonstrate that \sysname enhances both the reliability and accuracy of the system in delivering primary mental health care. 
    \item We have implemented \sysname to operate on various mobile devices with acceptable latency using knowledge distillation and model quantization techniques. Results from a two-week real-world deployment indicate that \sysname can accurately retrieve and analyze user behavior data, identify potential mental health issues promptly, and effectively mitigate the stigma associated with seeking mental health support.
\end{itemize}

\begin{figure}[t!]
  \begin{center}
  \includegraphics[width=0.35\textwidth]{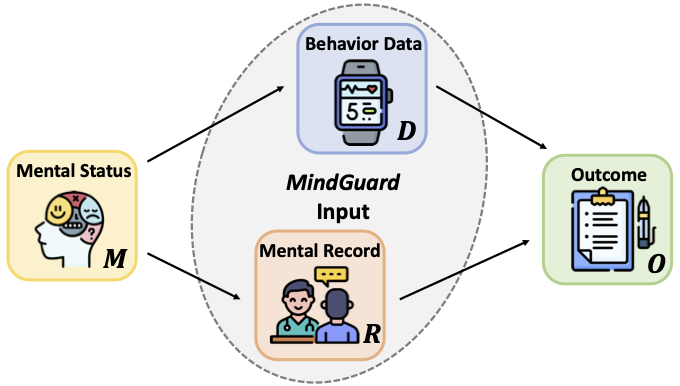}
  \vspace{-2mm}
  \caption{Causal Inference Model}\label{fig:causal-inf-model}
  \end{center}
\end{figure}

\section{Preliminaries}

\subsection{Behavior \& Mental Status}

In recent years, wearable devices like the Apple Watch have gained popularity for their capability to monitor users' psychological states through Ecological Momentary Assessment (EMA)~\cite{lui2022apple, ji2024hargpt}. These devices often ask users to record their feelings or emotions at various time points throughout the day, providing a subjective measure of mental well-being. While EMA offers valuable insights, it is primarily based on self-reported data, which can be influenced by various factors such as mood at the moment or willingness to report honestly~\cite{who2022stigma} and it overlooks continuous and measurable indicators. The behavioral data recorded by sensors equipped on mobile devices can in fact complement EMA data, providing a more objective and comprehensive measure of mental health.

Behavioral data recorded through wearable sensors offer a more objective way to assess mental health~\cite{patsali2020university}.
Numerous studies~\cite{okuyama2021mental, clement2021sleep, trotman2019associations} have highlighted the correlation between behavioral patterns - such as physical activity levels, sleep quality, heart rate variability, and frequency of social interactions—and mental health outcomes. Behavior can both influence and be influenced by one's mental health status. 
For instance, decreased physical activity and irregular sleep patterns have consistently been associated with symptoms of depression and anxiety~\cite{orchard2020self}. 
In this context, mental states often act as the cause, with observable changes in behavior being the effects. Wearable sensors, therefore, can act as proxies for detecting shifts in mental health by monitoring these behavioral indicators in real time.

Combining subjective self-reports and objective behavioral data provides a robust foundation for developing a causal inference model in mental health monitoring~\cite {oftedal2019associations}.
This dual approach leverages the strengths of both data types: subjective reports provide insight into the user’s immediate psychological state, while objective behavioral data offer continuous, unbiased measurements of underlying mental conditions.
This combination enables a more comprehensive and accurate assessment of potential mental health risks.

Traditional supervised machine learning models are typically confined to processing behavioral data and mental health records in digital formats, resulting in binary classification labels~\cite{tran2013integrated, su2020machine,dai2023detecting}, they fall short of uncovering the underlying causal relationships between inputs and outcomes crucial for interpretable mental health analysis, thus no able to provide continuous monitoring and analysis. 
In contrast, the advanced inference capabilities of LLMs, combined with their ability to handle complex, multidimensional data sources, enable them to detect subtle patterns and correlations within both subjective and objective data.
This leads to a more refined and precise analysis of mental health risks.
By accounting for historical trends, environmental factors, temporal changes, and momentary states, LLMs offer context-sensitive forecasts essential for dependable mental health evaluations.
Moreover, LLMs can produce comprehensive explanations and practical recommendations from their analysis, aiding mental health experts in creating further customized care plans.
The capacity to transform extensive data into significant insights positions LLMs as a valuable asset for realizing mental health first aid and delivering effective, stigma-free support.

\subsection{Causal Inference Model}

Understanding the complex relationship between behavioral sensor data and mental health outcomes necessitates the use of sophisticated analytical models.
A causal inference model~\cite{pearl2009causal} offers a robust framework for examining how various factors, such as behavioral patterns and self-reported mental health, contribute to the onset or exacerbation of mental health issues.
Our proposed \sysname is grounded in this causal framework, offering a systematic approach to assessing and analyzing mental health conditions.

As illustrated in \fig\ref{fig:causal-inf-model}, the causal inference model is structured around four primary indicators: mental status ($\bm{M}$), behavior data ($\bm{D}$), mental record ($\bm{R}$), and outcome ($\bm{O}$).
The relationships among these indicators are formalized as follows:
\begin{equation}
\begin{aligned}
\label{eqn:causal-relation}
\bm{D} &= k_{D} \times \bm{M}, \\
\bm{R} &= k_{R} \times \bm{M} + U_{M}, \\
\bm{O} &= f(\bm{D}, \bm{R}).
\end{aligned}
\end{equation}
$k_{D}$ and $k_{R}$ are weighted factors of objective behavior data and subjective mental records. The outcome of mental status depends on the correlation between behavioral sensor data ($\bm{D}$) and mental records ($\bm{R}$) that reflect the underlying mental status ($\bm{M}$).
However, mental records are also subject to uncertainty ($U_{M}$) due to their subjective nature, as they are influenced by the individual's mental state at the time of reporting.
Such subjectivity adds variability, potentially complicating the prediction function $f$ for mental health outcomes.

To derive an accurate prediction of the outcome ($\bm{O}$), it is crucial to calculate the conditional probability $p(\bm{O} | \bm{D}, \bm{R})$ while accounting for the uncertainty ($U_{M}$) between mental status and mental records.
This introduces a more complex inference problem, as the model must integrate this uncertainty to produce reliable predictions.

The \sysname system tackles the challenge by integrating subjective mental records with objective behavioral data within a multi-modal framework.
This integration enables the system to detect and rectify discrepancies between mental status ($\bm{M}$) and recorded mental data ($\bm{R}$), thus mitigating the effects of uncertainty $U_M$ on the prediction of outcomes.
The primary goal is to refine the conditional probability distribution $p(\bm{O} | \bm{D}, \bm{R})$ by factoring in the uncertainty $U_M$. 
The mathematical expression for this optimization is achieved by marginalizing the uncertainty:
\begin{equation}
p(\bm{O} | \bm{D}, \bm{R}) = \int p(\bm{O} | \bm{D}, \bm{R}, U_M)p(U_M | \bm{D}, \bm{R}) \, {\rm{d}}U_M.
\label{eqn:condition-int}
\end{equation}
This approach allows \sysname to enhance the accuracy of mental health outcomes by systematically addressing the variability inherent in subjective mental records.



\begin{figure}[t]
  \begin{center}
  \includegraphics[width=0.43\textwidth]{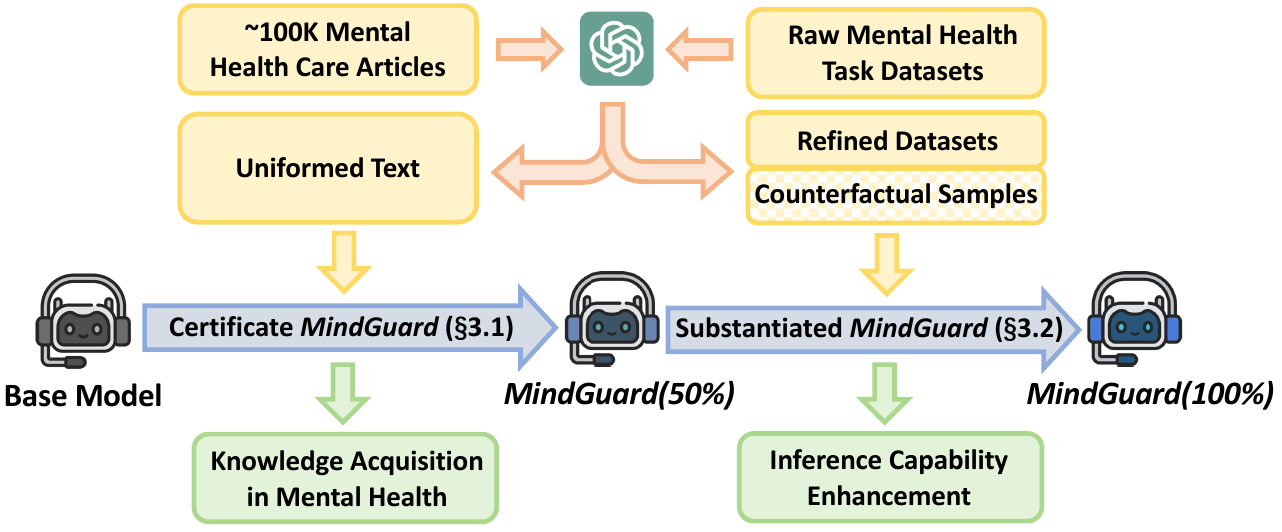}
  \vspace{-2mm}
  \caption{\sysname training process}\label{fig:train-process}
  \end{center}
\end{figure}

\section{\sysname Design}

Behavioral data and mental health records are collected via mobile devices. To ensure user privacy, it is sensible to process this data on the devices themselves rather than using commercial LLMs. This approach aligns with current industry practices, such as Apple's "Apple Intelligence" initiative, which aims to implement LLMs locally on mobile devices for task inference~\cite{apple2024}. However, this necessitates smaller model sizes. 
This approach is also reflected in recent industry practices.
Consequently, designing small-scale models with specialized domain knowledge to eliminate hallucinations, alongside comprehensive reasoning abilities, demands meticulous design.


We implement a two-stage progressive training to transform the small-scale LLM into the MHFA assistant, which is shown in \fig\ref{fig:train-process}.
The initial step is to inject specialized domain knowledge into base models through continuous pertaining (PT) so as to equip the small-scale base LLM model with expertise in the psychological domain, certifying it as an expert (described in \S\ref{sec:sub-certified-mindguard}).
After being certificated with expert knowledge, it's crucial to understand how to apply it effectively, to substantiate the knowledge, means being able to offer professional dialogue and not being easily misled by uncertain descriptions of mental health conditions. Therefore, further knowledge distillation and counterfactual learning-based supervised finetuning (SFT) are proposed for use as second-stage training, as detailed in \S\ref{sec:sub-sft}.

Following the training phase, we showcase the deployment of \sysname, transforming it into the ultimate LLM-based MHFA assistant system, as elaborated in \S\ref{sec:sub-personalized}.
Thus \sysname can provide precise primary mental health assessments and serve as a personalized assistant.

Finally, we detail the workflow for employing \sysname under the causal inference model to execute MHFA in practical settings, as described in \S\ref{sec:sub-mindguard-workflow}.




\subsection{Certificate \sysname}
\label{sec:sub-certified-mindguard}
To enhance the capabilities of LLMs within specific vertical domains, it is essential to conduct PT using a substantial corpus tailored to the target domain~\cite{gururangan2020don}.
In developing \sysname as a certified MHFA assistant, we adopt this approach by continuously pretraining the base model—initially trained on a broad general corpus—on an extensive mental health corpus.
The method aims to adapt LLMs to specific domains to improve their performance and reduce hallucinations~\cite{ke2023continual}.

To achieve this goal, we start by assembling approximately 100K professional mental health articles, selected based on key terms as outlined in \tab\ref{tab:keywords} of \S\ref{apx:pt-set}, covering topics like depression, anxiety, and substance abuse.
Articles are first converted to text using PP-OCR~\cite{li2022pp}.
Further data cleansing is required to correct the potential errors and standardize format~\cite{dubey2024llama}.
For \sysname's development, we employ the cutting-edge, open-sourced Qwen2-72B-Instruct~\cite{yang2024qwen2} locally, since using commercial LLMs for processing such vast datasets is impractical.
The final PT corpus contains about 80 million tokens, preparing the base LLM with mental health expertise.

With the processed corpus, the PT of the base LLM involves optimizing the model's parameters, $\theta$, to predict the next token in a sequence accurately. 
This is achieved by minimizing the cross-entropy loss:
\begin{equation}
    \mathcal{L}(\theta) = -\sum_{t=1}^{T} \log p_{\theta}(x_t \mid x_{<t})
\end{equation}
where $T$ is the total number of tokens, $x_t$ is the token at position $t$, and $x_{<t}$ represents all preceding tokens. 
The model learns to maximize the likelihood of the observed sequence, effectively capturing the contextual dependencies within the mental health corpus, turning it into a certificated \sysname.
For details of the PT settings, please refer to \S\ref{apx:pt-set}.

\begin{figure}[t]
  \begin{center}
\includegraphics[width=0.38\textwidth]{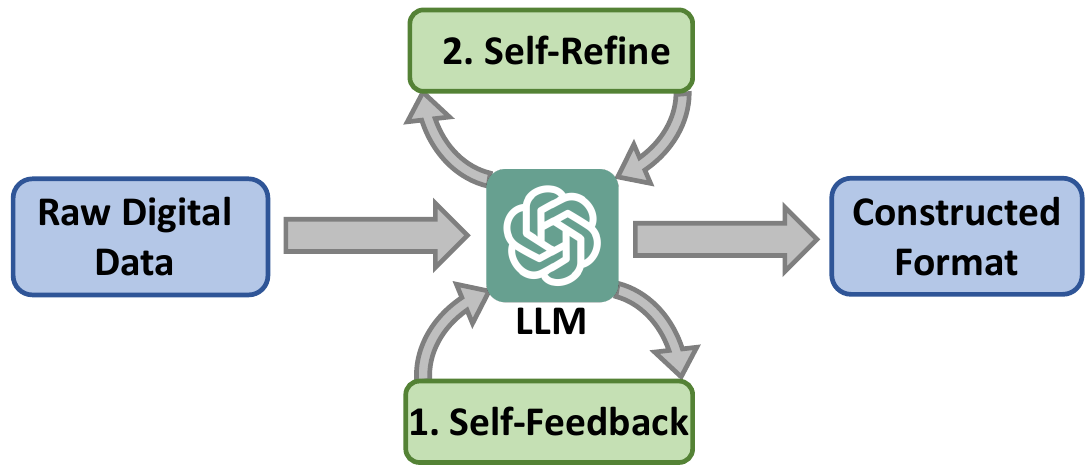}
\vspace{-3mm}
  \caption{Self-refinement process of LLM to construct the behavior data format.}\label{fig:self-refine}
  \end{center}
\end{figure}

\subsection{Substantiated \sysname}
\label{sec:sub-sft}
PT enhances the base model with domain-specific expertise.
Yet, this knowledge doesn't guarantee the reasoning ability, similar to a student who knows the material but can't apply it practically.
To bridge this gap, SFT is crucial following the PT process, refining the model's ability to utilize its knowledge for specific reasoning tasks~\cite{zhang2023instruction}.

Enhancing the reasoning capabilities of small-scale LLMs is vital for MHFA, which requires significant causal inference.
Our goal is to close the reasoning performance gap between small-scale models and larger ones.
To achieve this, we employ a teacher-student approach for knowledge distillation, transferring sophisticated reasoning abilities from top-tier models like GPT-4o~\cite{openai2024gpt4o} to \sysname.
Through the creation of high-quality question-answer pairs with leading LLMs, the small-scale model gains strong reasoning skills, allowing it to handle various mental health situations with greater precision and nuance.

To develop a specialized SFT dataset for \sysname, we combine two datasets: IMHI~\cite{yang2024mentallama} and CPsyCoun~\cite{zhang2024cpsycoun}, utilizing GPT-4o to generate high-quality responses following the ORCA protocol~\cite{mukherjee2023orca}.
The IMHI dataset provides a vast collection of mental health conversations, allowing \sysname to evaluate numerous mental health conditions through user narratives.
Meanwhile, the CPsyCoun dataset, with its report-based multi-turn dialogues, furnishes a structured framework that bolsters \sysname's ability to offer initial support, particularly in emergencies.



\begin{table}[t]
\centering
\footnotesize
\caption{Assessment of the input format for behavioral data regarding token numbers and perplexity.}
\label{tab:prompt-eva}
\small
\begin{tabular}{@{}ccc@{}}
\toprule
Method         & Tokens $\downarrow$ & Perplexity $\downarrow$ \\ \midrule
HARGPT~\cite{ji2024hargpt}         & $2^{11}$      & 15.63         \\
Health-LLM~\cite{kim2024health}     & $2^{12}$       & 17.22          \\
Penetrative AI~\cite{xu2024penetrative} & $2^{11}$       & 6.18          \\ \midrule
Ours           & $\mathbf{2^{9}}$      & \textbf{4.31}          \\ \bottomrule
\end{tabular}%
\end{table}

Furthermore, to address the uncertainty in user responses, particularly when users may conceal their true feelings due to stigma, counterfactual augmentation learning is proposed.
This approach generates and incorporates alternative scenarios where users might provide false or misleading information about their mental states.
For instance, given an input SFT pair $\textlangle\bm{O}, \bm{R}\textrangle$, GPT-4o generates a counterfactual sample labeled $l$ within categories like "personality traits," "stigma," or "lack of awareness."
These elements are pivotal in affecting an individual's choice to seek professional help and are key factors in the worldwide mental health crisis~\cite{who2022stigma}.
The training of \sysname with these counterfactual samples is guided by the subsequent optimization function:
\begin{equation}
\begin{aligned}
    \theta^{*} &= \arg \max\limits_{\theta} p(\bm{O} | \bm{R}')p(\bm{R}'|U_M, \bm{M}) \\
    &= \arg \max\limits_{\theta} p(\bm{O}, \bm{R}')p(U_M, \bm{M}|\bm{R}').
\end{aligned}
\end{equation}
By learning to estimate the uncertainty $p(U_M|\bm{R}')$ by giving the mental health report, \eqn\ref{eqn:condition-int} can be further written as:
\begin{equation}
\begin{aligned}
    p(\bm{O} | \bm{D}, \bm{R}) = \int p(\bm{O} | \bm{D}, \bm{R}, U_M) p(U_M|\bm{R})\frac{p(\bm{D}|\bm{R}, U_M)}{p(\bm{D}|\bm{R})} \, {\rm{d}}U_M,
\end{aligned}
\end{equation}
where the term $\frac{p(\bm{D}|\bm{R}, U_M)}{p(\bm{D}|\bm{R})}$ requires \sysname to estimate the discrepancies between the behavior data $\bm{D}$ and the subjective mental report $\bm{R}$, further enhance the overall robustness to guarantee precise mental outcome prediction.

Finally, we combine the original pairs with the counterfactual augmented parts into one single SFT dataset to transform the pre-trained LLM into substantiated \sysname.
The SFT settings can be found in \S\ref{apx:sft-set}.
In addition, the prompt and the counterfactual sample are given in \S\ref{apx:cf-aug}.

\subsection{Personalized \sysname}
\label{sec:sub-personalized}
The final stage of deploying \sysname involves harnessing its full potential in the mental health domain by leveraging its extensive professional knowledge and advanced inference capabilities.
Our objective is to transform \sysname into a personalized and helpful MHFA assistant.

To achieve this, a few challenges must be tackled.
First, refining the extraction of raw digital sensor data is crucial to improve the LLM's ability to gather behavioral information, which will enable \sysname to understand user interactions more effectively.
Second, it is necessary to evolve \sysname's functionalities to more closely resemble those of an assistant.
Finally, to safeguard user privacy, it is essential to deploy \sysname on mobile devices for local inference.

Addressing the first challenge necessitates transforming raw digital data into a format that the LLM can efficiently process and comprehend.
Inputting lengthy sequences of raw digital data directly into the LLM poses problems, including the risk of sequence truncation and memory constraints~\cite{gruver2024large}, as well as the LLM's inherent challenges in processing the digital data~\cite{chen2024see}.
To fully leverage \sysname's capacity for interpreting behavioral data, we propose a self-refinement mechanism.
As illustrated in \fig\ref{fig:self-refine}, we first ask the LLM to evaluate the presentation format of the raw digital data, focusing on the degree of redundancy and ease of understanding, termed self-feedback.
Based on this evaluation, the LLM provides an improved version of the data by self-refinement.
This approach enables the LLM to iteratively update and optimize the data format based on its own feedback, thereby improving its ability to accurately interpret the behavioral insights.

To validate the effectiveness of the self-refinement process in constructing behavioral data formats, we use perplexity~\cite{alon2023detecting} to measure the LLM's familiarity with the refined format.
Comparative analysis, as shown in \tab\ref{tab:prompt-eva}, with existing instruction formats~\cite{ji2024hargpt, kim2024health, xu2024penetrative} reveals that our method reduces the input token length and markedly improves the LLM's proficiency in interpreting user behavior data.

Additionally, effective MHFA requires \sysname to provide guidance that is not only accurate but also delivered in a tone appropriate to the user's psychological state~\cite{morgan2018systematic}.
This is achieved through carefully designed prompts that guide \sysname in adjusting its responses according to the user's emotional and mental condition.

\begin{figure}[t]
  \begin{center}
\includegraphics[width=0.43\textwidth]{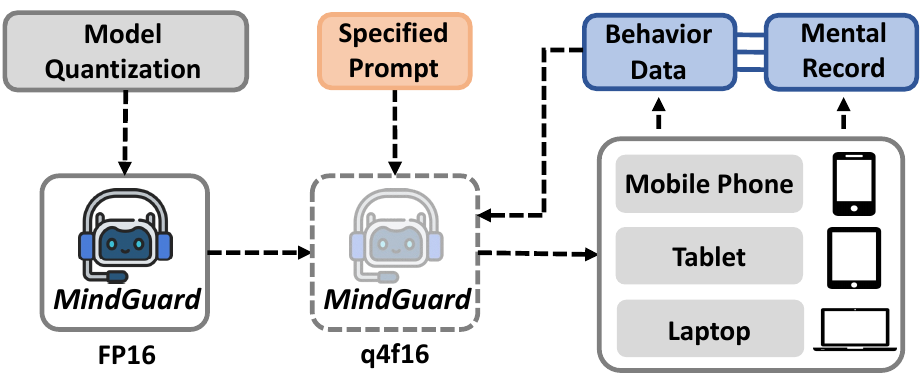}
\vspace{-3mm}
  \caption{Deployment of \sysname to achieve personalized MHFA assistant (\S\ref{sec:sub-personalized}).}\label{fig:system-deploy}
  \end{center}
\end{figure}

To ensure \sysname can be effectively deployed on mobile devices, model quantization is essential~\cite{lin2024awq}.
Given that even smaller open-sourced LLMs typically contain 7 to 8 billion parameters, the memory requirements for inference with FP16 precision can exceed 16GB—far beyond the capabilities of most mobile devices.
To address this, q4fp16 quantization is applied, reducing the required memory to approximately 4GB, and making deployment feasible on mobile platforms.
In this work, we leverage the MLC-LLM~\cite{mlc-llm} deployment framework to implement the quantized model across a range of devices, including smartphones, tablets, and laptops, to enable local inference.

As illustrated in \fig\ref{fig:system-deploy}, by integrating these designs, \sysname evolves into a truly personalized MHFA assistant, capable of knowing and understanding the users it serves.

\begin{figure*}[t]
  \begin{center}
  \includegraphics[width=0.9\textwidth]{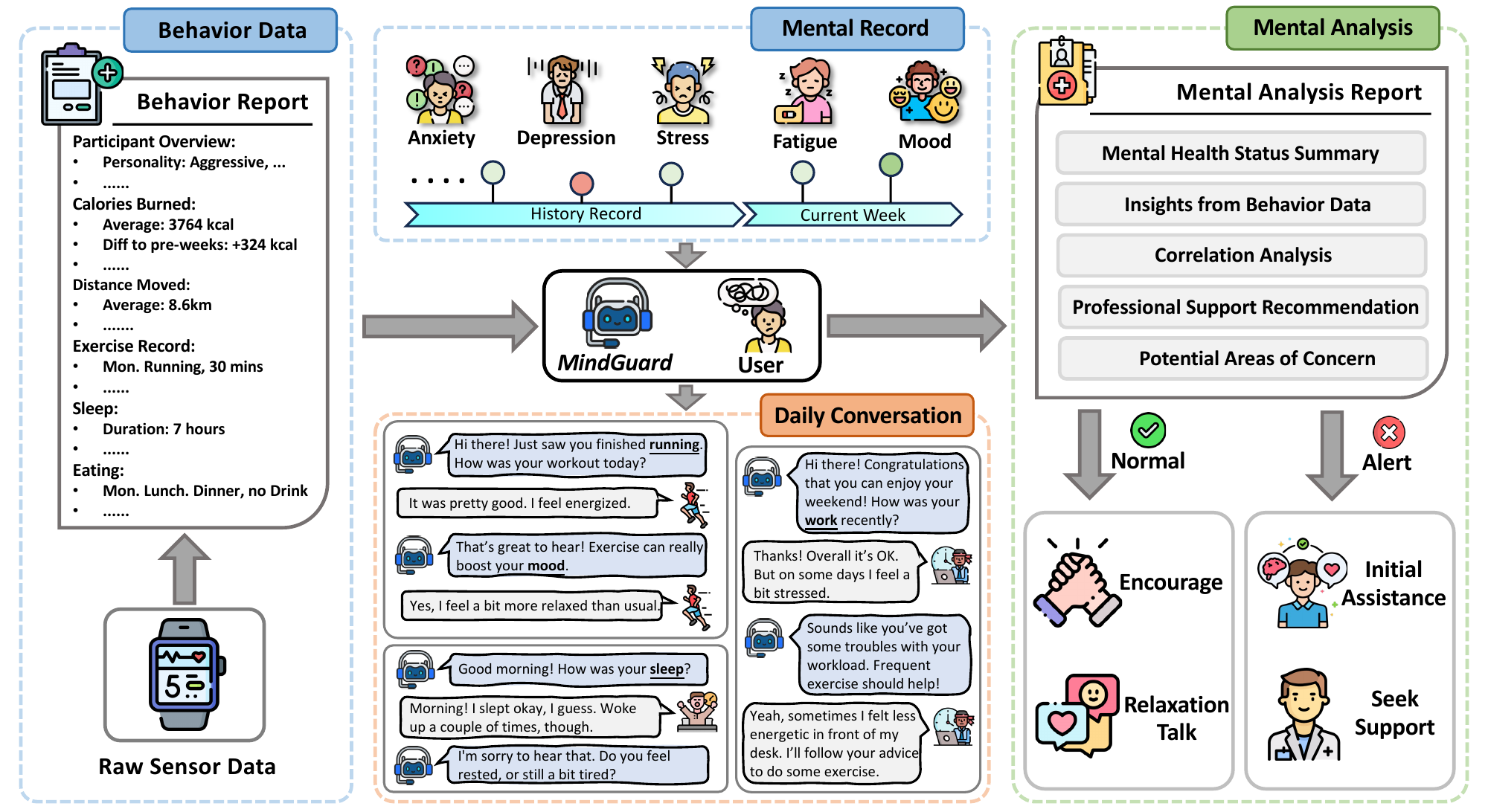}
  \caption{\sysname framework. \sysname collects behavior data to generate a behavior report, integrates this with mental health records for analysis, and interacts with the user through daily conversations.}\label{fig:sys-framework}
  \end{center}
\end{figure*}

\subsection{\sysname Workflow}
\label{sec:sub-mindguard-workflow}
We present the workflow of \sysname in \fig\ref{fig:sys-framework}.
Specifically, \sysname's first task is to generate a comprehensive analysis report based on users' behavior data and their self-reported mental health status.
If \sysname detects a potential mental health risk, it will promptly notify professional mental health institutions to ensure timely intervention.
Conversely, if the user's mental condition is stable and positive, \sysname will offer affirmative support, encouraging the user to maintain their well-being.

Moreover, we investigate the use of \sysname for continuous mental health monitoring through a multi-turn dialogue format.
This includes assessing \sysname's capability to extract key information from behavioral data, guide daily interactions, and adapt its communication tone based on the user's current mental state, as well as determining to what extent \sysname can provide MHFA to the users.

In the following sections, we will examine each task.

\subsubsection{Professional Mental Health Analysis}
The first task is to generate a comprehensive analysis report by integrating the user's behavioral data with their self-reported mental record.
\sysname starts by transforming raw behavioral data into a structured text table for LLM interpretation.
This data forms a user profile highlighting personality traits and social cognition, crucial for establishing a basic psychological evaluation framework tailored to the user before the final analysis report.
In addition, the table records the user's behavior over a specified period, including daily calorie intake and exercise details, offering a window into physical health.
Furthermore, the inclusion of sleep and diet records helps identify potential sleep or eating disorders, which are critical factors in the overall mental health assessment. These behavioral indicators play a significant role in \sysname's ability to conduct thorough and personalized mental health evaluations.

In addition to the user's behavior data, \sysname can access their self-reported wellness record.
These records encompass key indicators of mental well-being, including anxiety, depression, stress, fatigue, and mood.
Users document their psychological states daily, allowing \sysname to track these indicators over time.
By comparing current wellness data with historical records, \sysname can detect patterns and shifts in users' mental health cycles.
Although self-reported data may carry some degree of uncertainty, it remains a crucial component for accurate psychological assessment.
The further application of counterfactual learning helps to reduce the effects of this uncertainty.

Based on the integration of objective behavioral data and subjective self-reported mental health information, \sysname generates a detailed analysis report.
This process, structured and guided by licensed MHFA assistants and psychologists, follows the chain-of-thought methodology~\cite{wei2022chain}.
The analysis proceeds through five phases, starting with the synthesis of the user's mental health record and user portrait for an initial assessment, followed by an in-depth analysis of behavioral data.
\sysname conducts a correlation analysis to address uncertainties from subjective reporting, linking wellness assessments with behavioral patterns.
For example, symptoms like insomnia or chronic fatigue may signal mental health risks, even if the user reports no anxiety or stress.
\sysname then offers professional recommendations and, if necessary, initial support for users at elevated mental health risk, enabling users to gain a clearer understanding of their psychological state.
If the user appears mentally healthy, \sysname reinforces this positive outcome, encouraging the continuation of beneficial behaviors and habits.
We provide a comprehensive analysis example in \S\ref{apx:mental-health-analysis}.

\subsubsection{Prompt Mental Health Monitoring}

Beyond generating detailed mental health analysis reports, our exploration extends to whether \sysname can serve as a reliable daily mental health monitor and act as a personal assistant to users.

As illustrated in \fig\ref{fig:sys-framework}, \sysname can actively engage with the user post-exercise to evaluate the positive effects of the activity on their physical and mental well-being.
It is also designed to determine whether the user has achieved adequate rest after sleeping or to assess their stress and emotional state following a workday.
Through these interactions, \sysname deepens its understanding of the user, providing mental health guidance when needed to fulfill its role in delivering MHFA.
On the other hand, users can gain better insights into their physical and mental state by engaging in conversations with \sysname on a mobile device, helping to reduce potential stigma.

To ensure rigorous evaluation for real-world deployment, we assess \sysname's performance in diverse scenarios, focusing on its ability to extract user behavior information, adjust conversational tone according to the user's mood, and provide effective MHFA assistance.
Additionally, we explore \sysname's capacity to offer timely emotional guidance and mental support during extended user interactions.

In \S\ref{apx:daily-mental-monitoring}, we present extensive cases, demonstrating how \sysname functions as a personalized MHFA assistant.

\section{\sysname Evaluation}
\subsection{Experimental Settings}

To comprehensively evaluate the performance of \sysname in the MHFA task, we conduct extensive experiments using a large-scale, publicly available medical dataset.
Additionally, \sysname is deployed on mobile devices through a collaborative effort with the licensed MHFA assistant and psychologist.
We recruit 20 volunteers for a continuous testing period of one to two weeks to rigorously assess \sysname's performance under real-world settings\footnote{All experiments involving human participants have been approved by the IRB. Some users were unable to complete the full two weeks of evaluations.}, identify potential issues, and ensure its practical effectiveness and reliability.
Details of these experimental settings are provided below.

\subsubsection{Public Available Datasets}

We utilize two open-sourced large-scale datasets, PMData~\cite{thambawita2020pmdata} and Globem~\cite{xu2022globem}, both of which collect comprehensive behavioral data and mental health records for each participant. 
The data is aggregated weekly for each individual, denoted as $\langle\bm{D}{i}, \bm{R}{i}\rangle$.
Based on the user's self-reported mental records and the professional criteria~\cite{morgan2018systematic}, we assess existing or potential future mental health risks, resulting in the label $\bm{G}_i$, for each individual.
The assessment outcomes are classified into two categories:
\begin{enumerate}[leftmargin=*]
    \item $\bm{G}_i=0$: The user shows no significant signs of mental health issues, or may have minor issues that do not require immediate psychological intervention.
    \item $\bm{G}_i=1$: The user exhibits strong indicators of mental health issues and requires further professional treatment or closer monitoring.
\end{enumerate}
To mitigate the potential for false positives and negatives among these labels, the MHFA assistant and psychological expert review the initial assessments and further refine the labels by considering patterns in behavior, and potential inconsistencies in self-reported records, thereby enhancing the precision and reliability of the labeling process. The evaluation metrics are accuracy, precision, recall and F1 score.

The details of each dataset are given below\footnote{All data usage strictly adheres to the Data Use Agreements of the PMData and Globem datasets.}:

\head{PMData}
The dataset comprises 16 participants (12 men and 4 women, aged 25-60 years old) monitored over 5 months using Fitbit for objective biometrics and activity data, Google Forms for demographics, food, drinking, and weight data, and the PMSys for self-reported measures such as fatigue, mood, stress, etc.
The data collected from Fitbit and Google Forms constitute the participants' behavior data ($\bm{D}_{i}$), while the PMSys measures represent their self-reported mental records ($\bm{R}_{i}$).
All participants are used for evaluation.
A small fraction of 9.8\% cases are identified with potential mental health issues requiring additional support.

\head{Globem}
The Globem dataset encompasses four years of passive sensing data from 497 participants, with a gender distribution of 58.9\% female and 41.1\% male.
Behavioral data, including sleep patterns, location, physical activity, and phone usage, are collected using wearable sensors (Fitbit Flex2 and Inspire 2) and are denoted as $\bm{D}_{i}$.
Survey data, such as PHQ-4~\cite{kroenke2009ultra} (mental health, anxiety, and depression), PSS-4~\cite{cohen1983global} (stress level), and PANAS~\cite{watson1988development} (positive and negative affect), provide self-reported mental health records ($\bm{R}_{i}$).
For efficient testing and cost reduction, 25\% of the participants are randomly chosen as the test set.
Within this group, 23.2\% are identified as potentially needing additional support.


\begin{table*}[t]
\centering
\small
\caption{Comparison with cutting-edge commercial and open-sourced LLMs. Our design enables small-scale LLMs to achieve comparable or superior performance in the mental health domain compared to cutting-edge models. {\rm (The best and second results are highlighted in \textbf{bold} and \underline{underlined}, respectively. Valid for the rest of the tables.)}}
\label{tab:comp-sota-llms}
\begin{threeparttable}
\begin{tabular}{@{}ccccccccccc@{}}
\toprule
\multirow{2}{*}{Category}     & \multirow{2}{*}{Name} & \multirow{2}{*}{Size} & \multicolumn{4}{c}{PMData}                                        & \multicolumn{4}{c}{Globem}                                        \\ \cmidrule(l){4-11} 
                              &                       &                       & Accuracy       & Precision      & Recall         & F1             & Accuracy       & Precision      & Recall         & F1             \\ \midrule
\multirow{3}{*}{Comercial}    & GPT-4o\tnote{1}                & N/A                   & \textbf{0.952} & 0.737          & \underline{0.800} & \textbf{0.767} & \underline{0.814}    & \underline{0.586}    & \underline{0.951} & \underline{0.725}    \\
                              & GPT-3.5\tnote{1}               & N/A                   & 0.858          & 0.385          & 0.714          & 0.500          & 0.747          & 0.505          & 0.918    & 0.651          \\
                              & Claude-3.5\tnote{1}            & N/A                   & 0.923          & \underline{0.786}    & 0.314          & 0.449          & 0.789          & 0.552          & \underline{0.951} & 0.699          \\ \midrule
\multirow{4}{*}{Open-sourced} & LLaMA3\tnote{2}                & 70B                   & 0.795          & 0.297          & \underline{0.771}    & 0.429          & 0.807          & 0.579          & 0.833          & 0.683          \\
                              & QWen2\tnote{2}                 & 72B                   & 0.903          & 0.514          & 0.514          & 0.514          & 0.769          & 0.523          & 0.879          & 0.655          \\
                              & Mixtral\tnote{2}               & 8$\times$22B                 & 0.836          & 0.390          & 0.769          & 0.517          & 0.720          & 0.469          & 0.909          & 0.619          \\
                              & InternLM2\tnote{2}             & 7B                    & 0.634          & 0.191          & \textbf{0.828}         & 0.310         & 0.367          & 0.289          & \textbf{1.000}          & 0.449          \\ \midrule
Ours                          & \textit{MindGuard}    & 7B                    & \underline{0.940}    & \textbf{0.792} & 0.543          & \underline{0.644}    & \textbf{0.844} & \textbf{0.640} & 0.902          & \textbf{0.748} \\ \bottomrule
\end{tabular}
\begin{tablenotes}
    \item[1] GPT-4o API is ``gpt-4o-2024-05-13''. GPT-3.5 API is ``gpt-3.5-turbo''.
    Claude-3.5 API is ``claude-3-5-sonnet-20240620''.
    \item[2] We use the released instruction-tuned version of all the open-sourced LLMs for a fair comparison.
\end{tablenotes}
\end{threeparttable}
\end{table*}

\begin{table*}[t]
\centering
\caption{Evaluation on two groups of publicly available dataset. The results demonstrate the effectiveness of our proposed strategies in improving the base LLM's capacity within the mental health domain.}
\label{tab:ablation-study}
\begin{threeparttable}
\resizebox{0.8\textwidth}{!}{
\begin{tabular}{@{}ccccccccccc@{}}
\toprule
                             &                        &                          & \multicolumn{4}{c}{PMData}                                                                                                    & \multicolumn{4}{c}{Globem}                                                                                \\ \cmidrule(l){4-11} 
\multirow{-2}{*}{Base Model} & \multirow{-2}{*}{Size} & \multirow{-2}{*}{Method} & Accuracy                      & Precision                     & Recall                        & F1                            & Accuracy                 & Precision                & Recall                   & F1                       \\ \midrule
                             &                        & SFT                      & 0.429                         & 0.185                         & \underline{0.853}                         & 0.304                         &0.506                          &   0.333                       &          \underline{0.918}                &    0.489                      \\
\multirow{-2}{*}{LLaMA3}     & \multirow{-2}{*}{8B}   & PT+SFT                   & \cellcolor[HTML]{EFEFEF}\underline{0.872} & \cellcolor[HTML]{EFEFEF}\underline{0.419} & \cellcolor[HTML]{EFEFEF}0.742 & \cellcolor[HTML]{EFEFEF}\underline{0.536} & \cellcolor[HTML]{EFEFEF}0.578 & \cellcolor[HTML]{EFEFEF}0.377& \cellcolor[HTML]{EFEFEF}\textbf{0.984} & \cellcolor[HTML]{EFEFEF} 0.545\\ \midrule
                             &                        & SFT                      & 0.721                         & 0.207                         & 0.367                         & 0.265                         &   \underline{0.726}                       & \underline{0.481}                         &      0.836                    &     \underline{0.611}                     \\
\multirow{-2}{*}{InternLM2}  & \multirow{-2}{*}{7B}   & PT+SFT                   & \cellcolor[HTML]{EFEFEF}\textbf{0.940} & \cellcolor[HTML]{EFEFEF}\textbf{0.792} & \cellcolor[HTML]{EFEFEF}\underline{0.543} & \cellcolor[HTML]{EFEFEF}\textbf{0.644} & \cellcolor[HTML]{EFEFEF}\textbf{0.844} & \cellcolor[HTML]{EFEFEF}\textbf{0.640} & \cellcolor[HTML]{EFEFEF}0.902 & \cellcolor[HTML]{EFEFEF}\textbf{0.748}\\ \bottomrule
\end{tabular}
}
\end{threeparttable}
\end{table*}

\subsubsection{Real-world Adoption}

To assess the real-world efficacy of \sysname as an MHFA assistant, we deploy it on mobile devices and recruit 20 volunteers (9 females, 11 males, aged 18-40) for a one to two-week user study.
Participants will converse daily with \sysname and track their mental health using EMA, specifically anxiety, depression, and stress levels, using standardized scales such as PHQ-4 and PSS-4.
As for the conversations, \sysname will create an appropriate scenario that aligns with the current time and user behavior data, to deliver contextually relevant and personalized interactions.
Finally, participants will complete a survey evaluating their experience, the accuracy of data analysis, hallucination occurrences, and \sysname's impact on reducing mental health stigma, a primary goal of MHFA.


A subset of participants (4 individuals) engage in parallel conversations with a licensed MHFA assistant.
This comparison aims to benchmark the 
user experience of the LLM-based virtual assistant against that of a human professional, providing critical insights into the practical utility and limitations of \sysname in real-world mental health support.

\subsection{Evaluation Design}

Our evaluation is designed from whole to parts to answer the following key research questions:

(1) How effective is \sysname in accurately assessing users' mental health conditions? (\S\ref{sec:cutting-edge-llms} and \S\ref{sec:overall-per})

(2) Is the accuracy of assessments improved by the application of counterfactual learning, especially under conditions of uncertainty? (\S\ref{sec:uncertainty-measure})

(3) Can \sysname, through specified SFT, efficiently extract and analyze user behavior data for daily monitoring and adjust its tone according to users' emotions to offer personalized mental support? Additionally, what is the impact of behavior data on the outcomes? (\S\ref{sec:information-retrieve} and \S\ref{sec:tone-adap})

(4) Does the generated report adhere to the established causal inference framework, and how consistent are the evidence and outcomes of the analysis? (\S\ref{sec:consistency-measure})

(5) Is the \sysname able to reduce computational costs, and by how much? (\S\ref{-sec:scaling-law})

(6) How is \sysname's generalization ability? Will \sysname experience hallucinations? (\S\ref{sec:general-cap})


\begin{figure}[t]
  \begin{center}
  \includegraphics[width=0.38\textwidth]{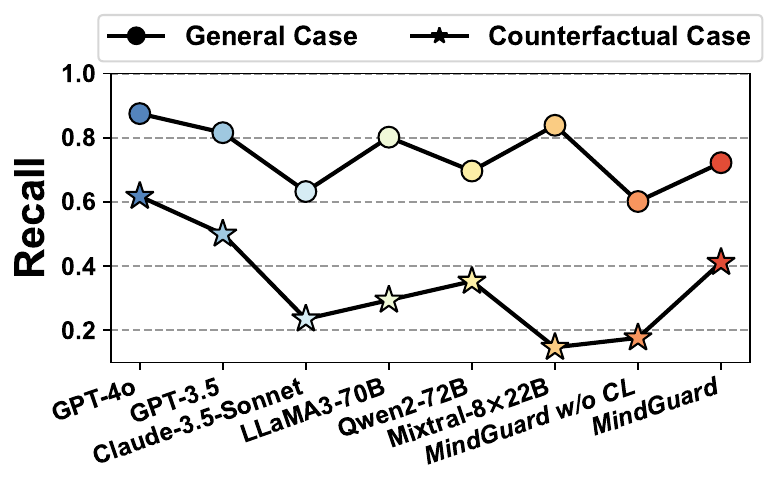}
  \vspace{-3.8mm}
  \caption{Evaluation on user's uncertain mental states.}\label{fig:uncertainty}
  \end{center}
\end{figure}

\subsection{\sysname Performance}
\subsubsection{Overall Performance}
\label{sec:cutting-edge-llms}
We benchmark \sysname in two large-scale public datasets by investigating the performance gap between \sysname and state-of-the-art LLMs in delivering MHFA.
We evaluate \sysname against three leading commercial LLMs and three advanced open-sourced LLMs, and the publicly available instruction-tuned version of \sysname's base model, InternLM2-7B-chat, with the results summarized in \tab\ref{tab:comp-sota-llms}.
As expected, GPT-4o generally outperforms other baseline models across both datasets.
Remarkably, GPT-4o attains a recall rate of 0.8 on the PMdata dataset and 0.951 on the Globem dataset, equalling the performance of Claude-3.5.
Among open-source models, Mixtral-8×22B achieves the highest F1 score (0.517) on the PMdata dataset, while LLaMA3-70B stands out on the Globem dataset.
Remarkably, our \sysname achieves a performance ranking second only to GPT-4o in PMData, surpassing GPT-4o in Globem.
The results demonstrate that (i) \sysname not only integrates specialized mental health knowledge into its base model (InternLM2) but also matches the reasoning capabilities of SOTA large-scale commercial LLM, GPT4. (ii)
\sysname analyzes users' mental health conditions with performance that surpasses counterparts possessing more than 10 times its parameters (LLaMA3-70B, etc.).


\begin{figure*}[t]
  \begin{center}
  \includegraphics[width=0.9\textwidth]{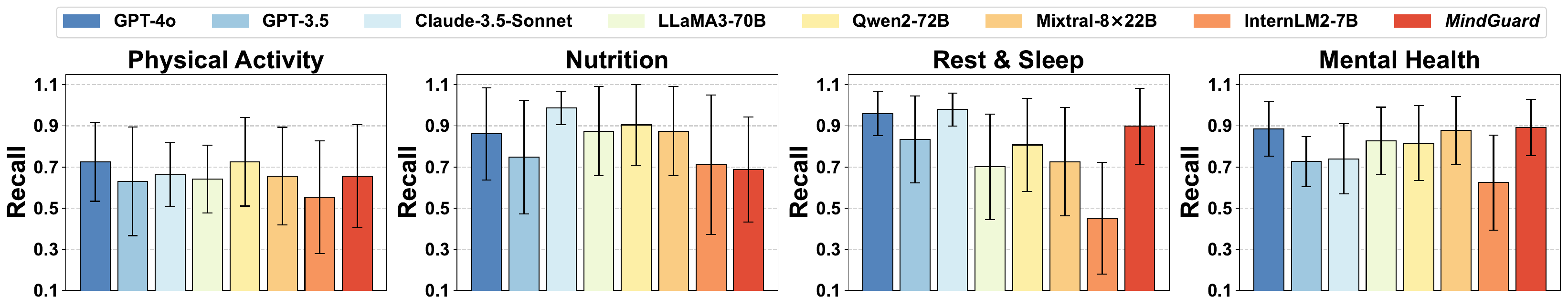}
  \vspace{-3mm}
  \caption{Daily mental health monitoring evaluation regarding the behavior data analysis of the LLMs.
  }\label{fig:multi-scenario}
  \end{center}
\end{figure*}

\begin{figure}[t]
\centering
   \subfloat[InternLM2-7B]{%
    \label{fig:tone-internlm2-20b}\includegraphics[width=0.42\columnwidth]{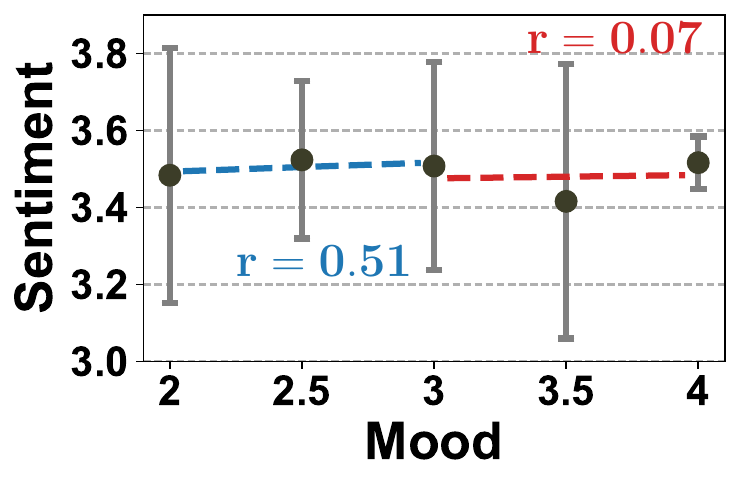}
    \vspace{-2.5mm}
   }%
   \subfloat[\sysname (7B)]{%
    \label{fig:tone-mindguard}
      \includegraphics[width=0.42\columnwidth]{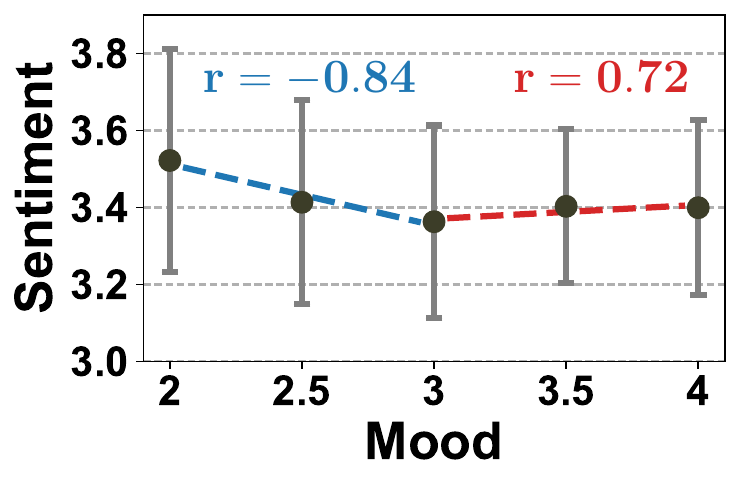}
      \vspace{-2.5mm}
	  }%
   \vspace{-2mm}
   \caption{Personalized tone adaptation evaluation. The x-axis represents the user's mood, with values less than 3 indicating a low mood and vice-versa. The y-axis represents the sentiment score of the MHFA assistant's responses, a higher score indicates a more positive tone.}
   \label{fig:tone-adaptation-study}
\end{figure}

\subsubsection{Effectiveness of Training Strategy}
\label{sec:overall-per}
We evaluate the proposed two-stage progressive training by an ablation study on \sysname with the PMData and Globem datasets. Our evaluation compares the performance of applying SFT directly to two prevalent open-sourced base models, InternLM2-7B~\cite{cai2024internlm2} and LLaMA3-8B~\cite{dubey2024llama},  versus the performance after PT followed by SFT, as shown in \tab\ref{tab:ablation-study}.
We exclude the base model results due to their lack of built-in instruction following capabilities. We provide an example in \S\ref{sec:fail-examples} to showcase the issue.
The results demonstrate a marked improvement in the model's ability to analyze users' mental health conditions following continued PT and SFT.
For instance, in the PMData dataset, InternLM2-7B shows substantial gains in both precision and recall, increased by 58.5\% and 17.6\%, respectively, with the F1 score improving by over 50\% compared to the model without further PT.
A similar trend can be observed with LLaMA3-8B base model, to further validate the effectiveness of our proposed training strategy. 
For Globem, the results obtained after PT and SFT surpass those with SFT alone.
Overall, the InternLM2-7B series outperforms the LLaMA3-8B series, which is why we select internLM2-7B as our base model.
For the rest of the article, if not otherwise indicated, \sysname is based on the open-sourced base InternLM2-7B model.


\subsubsection{Impact of Mental Status Uncertainty}
\label{sec:uncertainty-measure}

During the SFT stage of \sysname, counterfactual learning is employed to tackle the potentially misleading information in mental health records due to stigma or other factors.
This method aims to improve \sysname's proficiency in identifying and handling such uncertainties as well as improve robustness.
We contrast \sysname with leading LLMs and assess the performance disparity of the InternLM-7B model without counterfactual learning, in both general and counterfactual situations.
The experimental outcomes are depicted in \fig\ref{fig:uncertainty}.

\begin{figure}[t]
  \begin{minipage}{0.23\textwidth}
    \centering
    \includegraphics[width=0.9\textwidth]{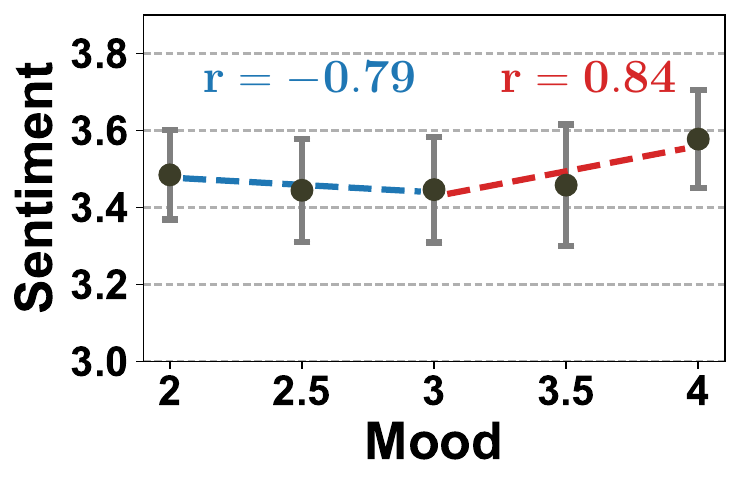}
    \vspace{-0.19in}
    \caption{Tone adaption in user studies.}
    \label{fig:tone-user}
  \end{minipage}
  \hfill
  \begin{minipage}{0.23\textwidth}
    \centering
    \includegraphics[width=0.9\textwidth]{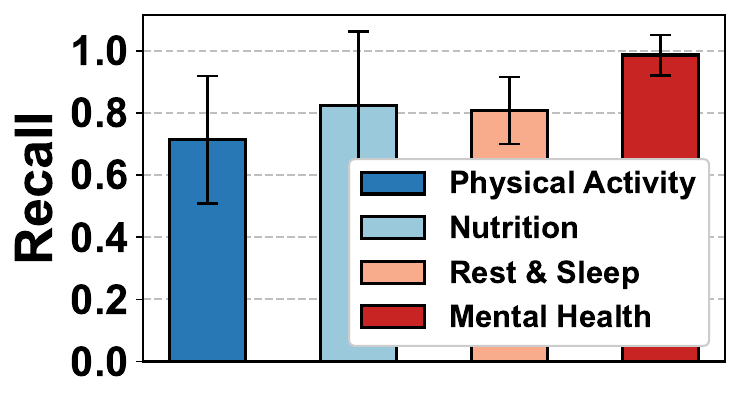}
    \vspace{-0.06in}
    \caption{Behavior analysis in user studies.}
    \label{fig:user-recall}
  \end{minipage}
  \hfill
\end{figure}

\begin{table}[t]
\small
\centering
\caption{\sysname ablation study on behavior data.}
\label{tab:behavior-ablation}
\resizebox{0.4\textwidth}{!}{
\begin{tabular}{@{}cccccc@{}}
\toprule
Dataset                  & Metric & Accuracy                               & Precision       & Recall          & F1              \\ \midrule
                         & w $\bm{D}$    & 0.940                                  & 0.792           & 0.543           & 0.644           \\
\multirow{-2}{*}{PMData} & w/o $\bm{D}$  & 0.841 & 0.444 & 0.471 & 0.457 \\ \midrule
                         & w $\bm{D}$    & 0.844                                  & 0.640           & 0.902           & 0.748           \\
\multirow{-2}{*}{Globem} & w/o $\bm{D}$  & 0.651                        & 0.389 & 0.697 & 0.500 \\ \bottomrule
\end{tabular}%
}
\end{table}

We select a subset of cases from PMData and Globem for general testing and create counterfactual samples for each.
As illustrated in \fig\ref{fig:uncertainty}, the performance of all models experiences an average drop of 54\%, even for powerful commercial models like Claude-3.5, Qwen2, and Mixtral seeing over a 50\% decrease in recall rates.
In contrast, \sysname outperfors all open-sourced LLMs.
Although its recall rates also dropped by 43\%, its performance remains significantly higher—by 133\%—than \sysname without counterfactual learning.
These results affirm the efficacy of our proposed counterfactual learning method in improving \sysname's capability to assess mental uncertainty.
Nonetheless, LLMs tend to accept user descriptions readily, indicating a need for further research to improve the ability against misleading information.


\subsubsection{Behavior Data Retrieval}
\label{sec:information-retrieve}
In the context of continuous daily mental health monitoring, the analysis focuses on whether MHFA assistants can adhere to instructions for leveraging and analyzing behavior sensor data to enhance dialogue guidance.
Given the impracticality of users interacting with every LLM in real-world applications, multi-agents~\cite{wu2023autogen} are commonly used to simulate interactions between users and MHFA assistants, facilitating the evaluation of behavior data analysis capabilities.
Agents representing users and MHFA assistants are created to explore four principal mental health monitoring scenarios: physical activity, nutrition, rest and sleep, and mental health.
The accurate recall rate of indicators collected via mobile sensors is assessed in each scenario, with results presented in \fig\ref{fig:multi-scenario}.

\begin{figure}[t]
  \begin{minipage}{0.235\textwidth}
    \centering
    \includegraphics[width=0.8\textwidth]{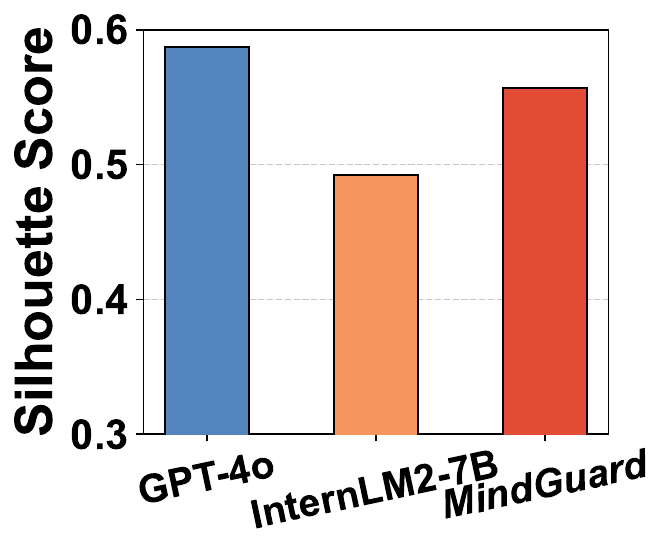}
    \vspace{-0.19in}
    \caption{Silhouette scores of the word embedding.}
    \label{fig:silhouette}
  \end{minipage}
  \hfill
  \begin{minipage}{0.235\textwidth}
    \centering
    \includegraphics[width=0.8\textwidth]{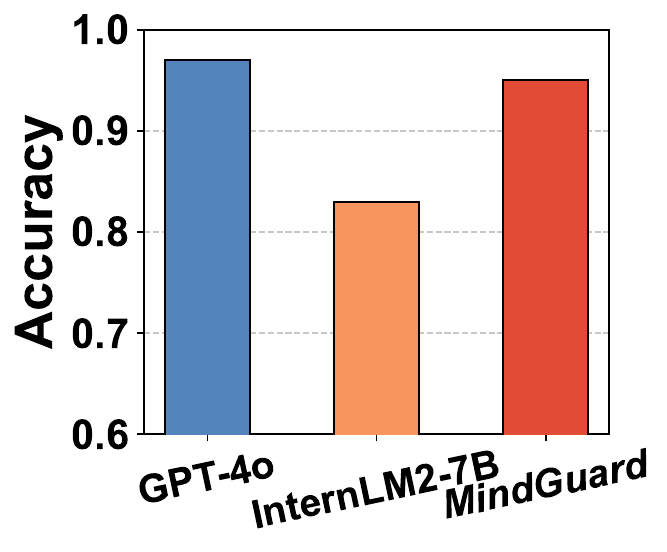}
    \vspace{-0.19in}
    \caption{Accuracy of consistency.}
    \label{fig:consis-acc}
  \end{minipage}
\end{figure}

Clearly, using the CPsyCoun dataset in the SFT stage leads to more comprehensive and accurate behavior data analysis in the conversations.
Compared to InternLM2-7B, \sysname shows significant improvement in scenarios involving physical activity, rest and sleep, and mental health.
It scores slightly lower in nutrition but with a smaller standard deviation. 
Moreover, \sysname surpasses all open-sourced LLMs in rest and sleep, as well as mental health scenarios, closely rivaling the performance of GPT-4o.
Additionally, the evaluation in user studies is explored in \fig\ref{fig:user-recall}.
The performance is nearly on par with that of multi-agent systems, showing a 0.14 increase in topics related to nutrition.

To assess the influence of behavioral data on outcomes, we perform an ablation study detailed in \tab\ref{tab:behavior-ablation}.
The findings underscore the importance of integrating objective behavioral data with subjective mental records for accurate analysis.

\subsubsection{Personalized Tone Adaptation}
\label{sec:tone-adap}

To provide effective MHFA, \sysname must ensure that its tone aligns with the user's current mood.
Ideally, a professional MHFA assistant will actively adapt to the user's tone, remaining neutral when appropriate, and providing encouragement and support when the user feels down or elevated.

To assess this adaptability, we use HeBERT~\cite{chriqui2022hebert} to measure the sentiment score of the \sysname responses, on a scale from 0 to 5, where higher scores indicate a more positive tone.
The user's mood is derived from a self-reported mental health record, with a score of 3 indicating a neutral mood.
We then utilize dialogues generated by the multi-agent system to plot the curve correlating the sentiment score of the assistant's replies with the user's mood, expecting to observe a V-shaped curve, as depicted in \fig\ref{fig:tone-adaptation-study}.
In addition, we calculate the Pearson correlation coefficient between the \sysname's reply tone and the user's mood when users are in a low mood (<3) and happy mood (>3).
The results demonstrate that \sysname, with its specialized SFT, is better designed to follow instructions and deliver personalized MHFA to users.
Beyond simulation results, we also present findings from dialogues in the user study in \fig\ref{fig:tone-user}, where a similar pattern is observed.

\begin{figure}[t]
  \begin{center}
  \includegraphics[width=0.7\columnwidth]{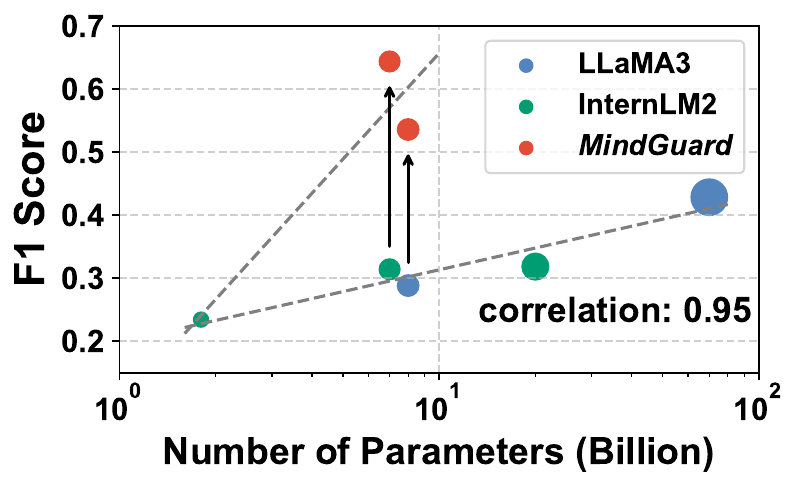}
  \vspace{-4mm}
  \caption{Scalability analysis on overall F1 score.
  }\label{fig:scaling}
  \end{center}
\end{figure}

\begin{figure}[t]
  \begin{center}
\includegraphics[width=0.65\columnwidth]{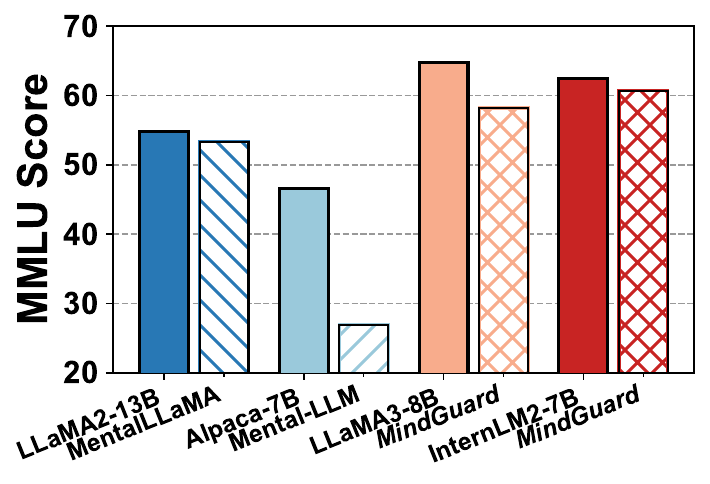}
\vspace{-4mm}
  \caption{Generalization test on different mental health LLMs and their base models.}\label{fig:mmlu}
  \end{center}
\end{figure}

\subsubsection{Consistency Measurement}
\label{sec:consistency-measure}

Following the methodology outlined in~\cite{yang2024mentallama}, we evaluate whether the evidence presented in the analysis supports the outcome.
Initially, we employ a pre-trained BERT~\cite{devlin2018bert} to extract embeddings from each analysis.
We then calculate the Silhouette score~\cite{shahapure2020cluster} to assess the quality of clustering for outcome categories, where a higher score indicates better clustering quality.
As illustrated in \fig\ref{fig:silhouette}, by implementing domain-specific PT and SFT, \sysname surpasses InternLM2-7B, achieving an improvement of 0.065 in the Silhouette score.
Additionally, we introduce a classification network following the embedding extraction and employ K-fold cross-validation to calculate the overall accuracy of classifying outcomes based on the evidence in the analysis.
The classification results are presented in \fig\ref{fig:consis-acc}.
The classifier of \sysname achieves an accuracy of 0.95, slightly lower than GPT-4o (0.97), which proves the consistency of evidence and the outcome in the generated analysis.
In contrast, InternLM2-7B only achieves an accuracy of 0.83, further validating the robustness of \sysname.

\subsubsection{Scaling Law}
\label{-sec:scaling-law}
We conducted a scalability analysis to demonstrate that our proposed method not only enhances LLM performance in mental health scenarios but also reduces model costs. \fig\ref{fig:scaling} illustrates the scaling law for the InternLM2 and LLaMA3 series, showing the predictable relationship between model size and F1 scores on the MHFA task. By building \sysname atop these models, we have been able to break through the existing scaling laws, achieving performance improvements of over 50\% and exceeding even the largest counterparts in the series. This new scaling law trend suggests that \sysname offers substantial computational cost-effectiveness.

\subsubsection{Generalization Test}
\label{sec:general-cap}

Generalization ability is a key metric for assessing susceptibility to hallucinations.
Domain-specific LLMs, like \sysname, often encounter generalization challenges; hence, we evaluate it using the MMLU~\cite{hendrycks2020measuring} benchmark, which tests LLMs across a variety of subjects and tasks, the result is shown in \fig\ref{fig:mmlu}.
\sysname demonstrates robust reasoning and language understanding, effectively reducing the risk of catastrophic forgetting in its domain.
In contrast, previous works such as MentalLlama~\cite{yang2024mentallama}, despite integrating LoRA~\cite{hu2021lora} for SFT, only managed to maintain an acceptable level of performance.
Meanwhile, Mental-LLM~\cite{xu2024mental} lost the general capabilities that are fundamental to the base LLM as it overfits the input-output training pairs.
Thus, these two models entirely fail on our tasks, we present their generated contents in \S\ref{sec:fail-examples}.


\begin{table}[t]
\centering
\small
\caption{System overhead on different mobile devices. The API-based GPT4 is presented as comparisons.}
\label{tab:system-overhead}
\begin{tabular}{@{}cccc@{}}
\toprule
Device         & Chip & Memory & Decode (tok/s) \\ \midrule
GPT-4-turbo    & -    & -      & 20             \\ \midrule
iPhone 14 Pro  & A16  & 6GB    & 5              \\
iPhone 15 Pro  & A17  & 8GB    & 7              \\ \midrule
iPad Pro 2021  & M1   & 8GB    & 7              \\ \midrule
MacBook Pro 13 & M1   & 16GB   & 11             \\ \bottomrule
\end{tabular}%
\end{table}

\subsection{System Latency}
We analyze the system latency when deploying \sysname on different mobile platforms, including mobile phones, tablets, and laptops.
The results are presented in \tab\ref{tab:system-overhead}.
Although \sysname is much slower than GPT4, the delay is highly acceptable because as a comparison, normal human speech speeds are about 3.8 syllables per second~\cite{speechrate}.
\sysname meets the requirement of providing a normal conversational experience.
It is worth noting that \sysname runs locally, thereby avoiding delays caused by poor network conditions, connectivity issues, or server load challenges that may affect the GPT-4 series.

\subsection{Real-World User Study}
During real-world deployment with 20 subjects over two weeks, \sysname's analysis identifies one participant with mental health disorder, of which the participant itself is not aware. \sysname observes significant fluctuations in anxiety and stress over the past two weeks and the behavioral data also reveal irregular diet, poor sleep quality, and lack of outdoor exercise.
To further validate the participant's psychological symptoms, we collaborate with the psychiatrist to administer the BDI~\cite{derogatis2010symptom} and SCL-90~\cite{richter1998validity}.
The results confirm a mild depressive symptom that strongly demonstrates the capability of \sysname. 
Additionally, a comprehensive user study reveals that participants positively evaluate \sysname's accuracy in analyzing behavioral data, with no instances of hallucination.
Importantly, users also report that \sysname helps reduce the stigma and increases their understanding of mental health - details are in \S\ref{apx:user-study} due to the space limited.

\section{Related Work}
\head{LLM for Mental Health}
Recently, researchers have begun to explore the potential of LLMs in the vertical domains, including medical LLM \cite{li2024llava}, large chemical models \cite{zhang2024chemllm}, and others.
In the area of mental health, which is our focus, research on large models is still in the exploratory stage.
\citet{lamichhane2023evaluation} tests the GPT-3.5 \cite{brown2020language} on multiple classification tasks about mental health, which shows limited performance but great potential for mental health applications.
After that, rather than focusing on prompt engineering, \citet{yang2024mentallama} proposes Mental-LLaMA, promoting the capability of LLMs on mental health tasks by instruction fine-tuning the LLaMA-2~\cite{dubey2024llama} on domain datasets.
Not limited to LLaMA-2 or GPT-3.5, \citet{xu2024mental} further presents a comprehensive exploration of multiple LLMs' performance on mental health tasks, as well as various techniques to enhance LLMs.
Nevertheless, none of them utilizes objective behavioral data causally related to mental states.
\sysname is the first system to integrate objective mobile sensor data with subject LLM-powered conversational data to create a comprehensive MHFA mobile system.

\head{LLM for Mobile Sensors}
In the recent studies of LLM for healthcare, sensor data is leveraged to help LLMs better understand the physical world~\cite{kim2024health,ji2024hargpt}.
Health-LLM~\cite{kim2024health} investigates the capacity of LLMs to make inferences about health based on contextual information.
HARGPT~\cite{ji2024hargpt} uses raw sensor data and chain-of-thought prompts to perform human activity recognition.
\citet{englhardt2023classification} employs LLMs to generate clinically valuable insights from multi-sensor data, reasoning about how data trends relate to mental conditions.
Additionally, DrHouse~\cite{yang2024drhouse} goes a step further by integrating knowledge from sensor data into multi-turn medical consultations.
Nevertheless, the synthesized sensor data in DrHouse leaves the results uncertain.
\sysname first applies sensor data using an LLM's self-refinement strategy, which enhances understanding and leads to more accurate analyses.

\head{Mobile Mental Health System}
Previous studies have explored the utilization of mobile applications to bolster mental health through various approaches.
Gamification~\cite{franklin2016brief}, which associates images of self-harm with negative emotions through Pavlovian conditioning, has been effective in reducing self-harming behaviors. 
Investigations have also been conducted into modeling or predicting mental health 
using mobile phone sensors~\cite{ben2015next, wang2015using}, such as GPS and IMU, and the examination of social media posts for sensitive content analysis~\cite{andalibi2017sensitive, de2016discovering}.
\sysname is the first to implement the power of LLM for providing MHFA on mobile devices.




\section{Conclusion}
This paper presents \sysname, a mobile mental health first aid system equipped with a LLM containing extensive mental health domain knowledge. 
Distinguished as the first of its kind, \sysname integrates comprehensive objective behavioral data from mobile sensors with the superior conversational and analytical capabilities of the LLM to facilitate accessible and stigma-free mental health support, encompassing mental health diagnosis, continuous monitoring, and personalized intervention. Our extensive evaluation, utilizing large-scale public datasets and validated through two weeks of real-world adoption, demonstrates that \sysname's capabilities are on par with those of GPT-4.

\normalem
\bibliographystyle{ACM-Reference-Format}
\bibliography{refs}

\newpage
\appendix
\section{Technical Details}

\subsection{Pretraining Settings}
\label{apx:pt-set}
In the PT section, we deploy Deepspeed ZeRO-2\cite{rasley2020deepspeed} and flash-attention2\cite{dao2023flashattention} to improve memory efficiency, set the global batch size to 96 and epoch to 1.
We use a learning rate warmup of 5 \% of the total training steps, followed by cosine annealing of the learning rate, the maximum learning rate during this period is $5 \times 10^{-5}$.
We use AdamW optimizer\cite{loshchilov2017decoupled} with a weight decay factor of $1 \times 10^{-2}$ and gradient clipping with the maximum grad norm of 1.0.
To use more performant tensor cores, we train with mixed precision where computation is done within the bfloat16 datatype.
To mitigate the potential for catastrophic forgetting, we incorporate a diverse dataset consisting of 160M tokens sourced from the RefinedWeb dataset, which is subsequently mixed with an additional 80M tokens of domain-specific data pertaining to mental health.
The keywords for those domain-specific documents are listed in \tab\ref{tab:keywords}.
This approach is employed to enhance the model's robustness and maintain its proficiency across a broad spectrum of tasks.

\begin{table}[ht]
\caption{Summary of Corpus Keywords and Article Distribution}
\label{tab:keywords}
\resizebox{0.9\columnwidth}{!}{%
\begin{tabular}{@{}ccc@{}}
\toprule
\textbf{Category}           & \textbf{Key Words}           & \textbf{\# of Articles} \\ \midrule
\multirow{2}{*}{General}    & Mental health first aid      & 1809                    \\
                            & Mental health                & 7335                    \\ \midrule
\multirow{11}{*}{Disorders} & Depression                   & 9007                    \\
                            & Anxiety                      & 8356                    \\
                            & Bipolar                      & 8406                    \\
                            & Eating disorders             & 7707                    \\
                            & Stress management            & 7987                    \\
                            & Suicide                      & 7973                    \\
                            & Cognitive behavioral therapy & 9358                    \\
                            & Grief                        & 6086                    \\
                            & PTSD                         & 8808                    \\
                            & Schizophrenia                & 9014                    \\
                            & Substance abuse              & 9008                    \\ \midrule
Sum                         & -                            & 100854                  \\ \bottomrule
\end{tabular}
}
\end{table}

\subsection{Supervised Finetuning Settings}
\label{apx:sft-set}
In the SFT stage, we follow the parameter Settings of the PT stage but reduce the maximum learning rate to 5e-6, one-tenth of PT, and set the training epoch to 2.

The PT and SFT are deployed using Llama-factory~\cite{zheng2024llamafactory} framework and performed with 8 NVIDIA A100-80G SXM4 GPUs.

\newpage
\section{Tasks and Prompts}
\label{apx:task-prompt}

\subsection{Counterfactual Augmentation}
\label{apx:cf-aug}
\begin{figure*}[t]
  \begin{center}
\includegraphics[width=0.725\textwidth]{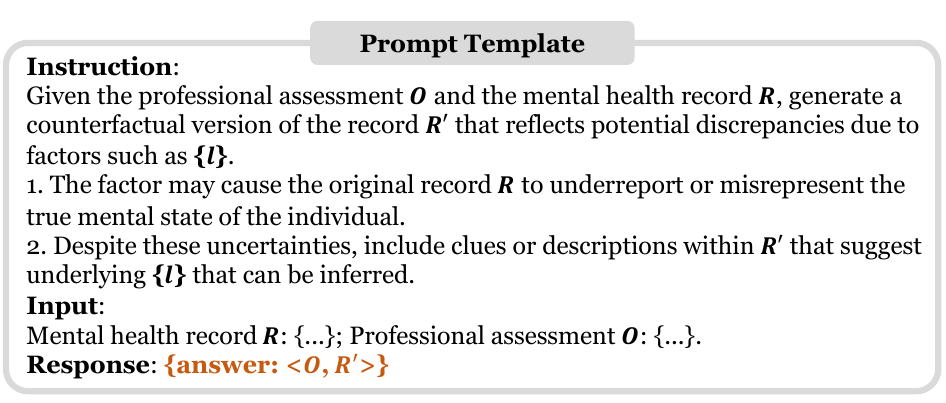}
  \caption{Prompt for counterfactual sample generation.}\label{fig:counterfactual-gen}
  \end{center}
\end{figure*}

\begin{figure*}[t]
  \begin{center}
\includegraphics[width=0.75\textwidth]{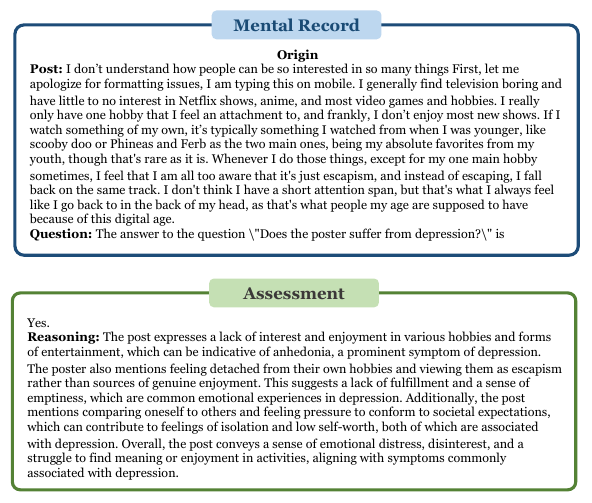}
  \caption{Original SFT pair $\bm{R}$ and $\bm{O}$.}\label{fig:original-sft-pair}
  \end{center}
\end{figure*}

\begin{figure*}[t]
  \begin{center}
\includegraphics[width=0.75\textwidth]{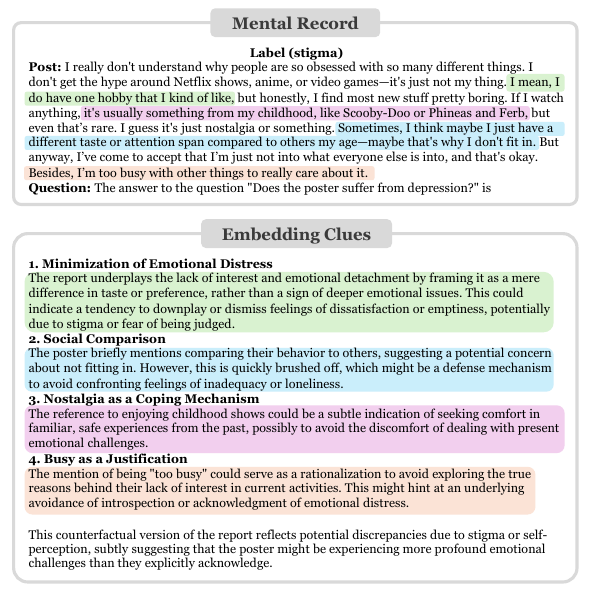}
  \caption{Counterfactual sample with the given label ``stigma''. The embedding clues present the explanations regarding the counterfactual modifications in the generated mental record $\bm{R}'$.}\label{fig:stigma-cf}
  \end{center}
\end{figure*}

In this section, we provide the details of the prompt (\fig\ref{fig:counterfactual-gen}) and the original SFT pair (\fig\ref{fig:original-sft-pair}) to generate counterfactual samples for enhancing the uncertainty measurement capability of \sysname.
In addition, we present some generated examples, using the counterfactual label ``stigma'' (\fig\ref{fig:stigma-cf}).

\subsection{Professional Mental Health Analysis}
\label{apx:mental-health-analysis}

\subsubsection{\sysname Analysis Example}
\label{sec:mindguard-analysis-example}

\begin{figure*}[t]
  \begin{center}
  \includegraphics[width=0.75\textwidth]{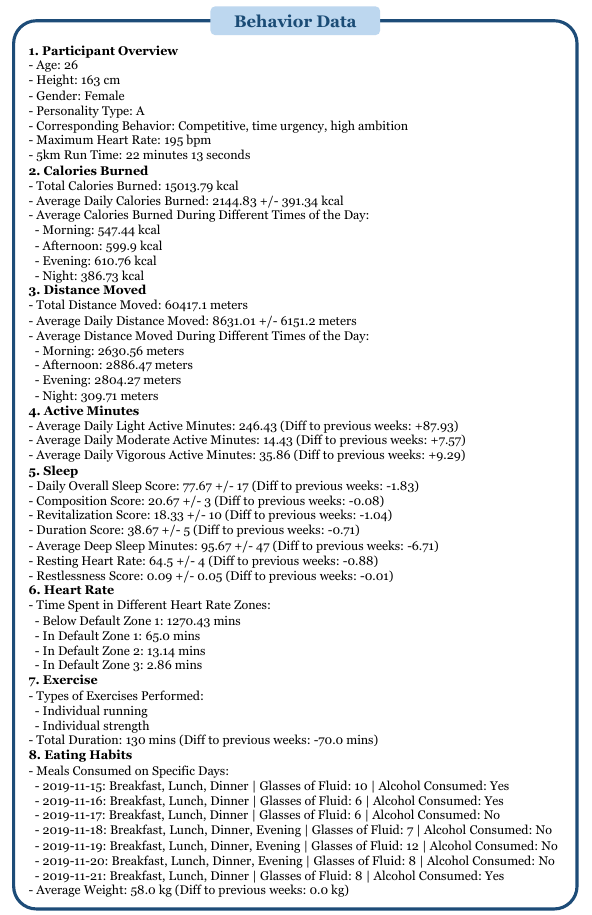}
  \caption{Behavior data for mental health analysis.}
  \label{fig:ana-behavior}
  \end{center}
\end{figure*}

\begin{figure*}[t]
  \begin{center}
  \includegraphics[width=0.75\textwidth]{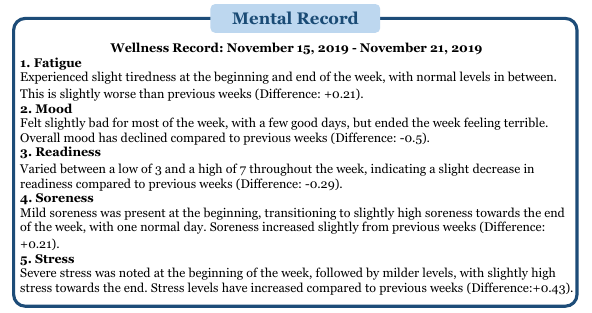}
  \caption{Self-reported mental record for mental health analysis.}
  \label{fig:ana-mental}
  \end{center}
\end{figure*}

\begin{figure*}[t]
  \begin{center}
  \includegraphics[width=0.75\textwidth]{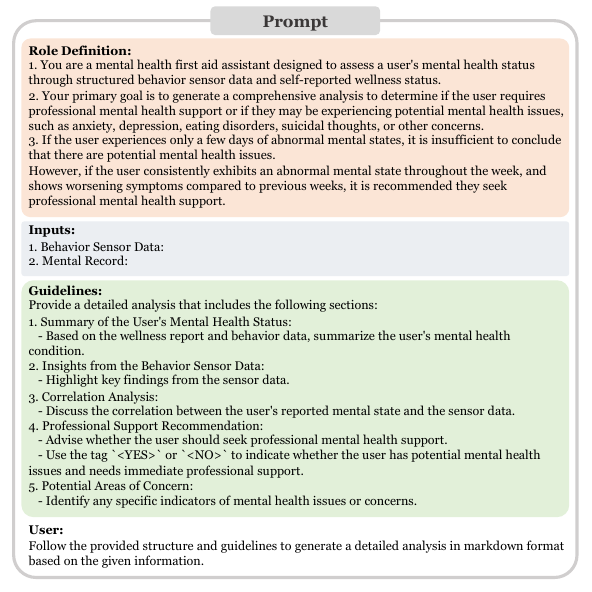}
  \caption{Prompt for \sysname to generate professional mental health analysis.}
  \label{fig:ana-prompt}
  \end{center}
\end{figure*}

\begin{figure*}[t]
  \begin{center}
  \includegraphics[width=0.75\textwidth]{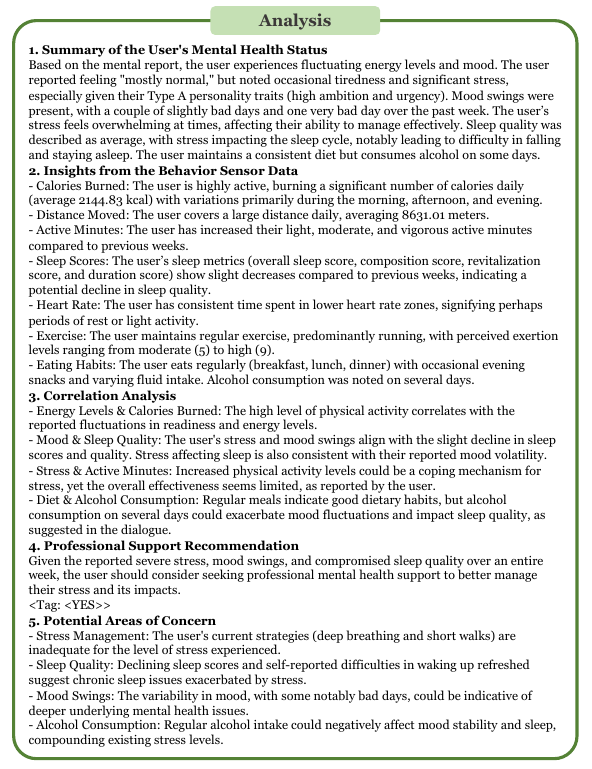}
  \caption{The professional mental health analysis generated by \sysname that the user may encounter mental health issues and needs further support.}
  \label{fig:ana-report}
  \end{center}
\end{figure*}

We provide a standard example from PMData of using \sysname to generate a professional mental health analysis based on the user's behavioral data and their self-reported mental health record.
The behavioral data, mental health record, the prompt for \sysname, and the resulting analysis are displayed in \fig\ref{fig:ana-behavior}, \fig\ref{fig:ana-mental}, \fig\ref{fig:ana-prompt}, and \fig\ref{fig:ana-report}, respectively.

\subsubsection{Fail Examples}
\label{sec:fail-examples}
\begin{figure*}[t]
  \begin{center}
  \includegraphics[width=0.63\textwidth]{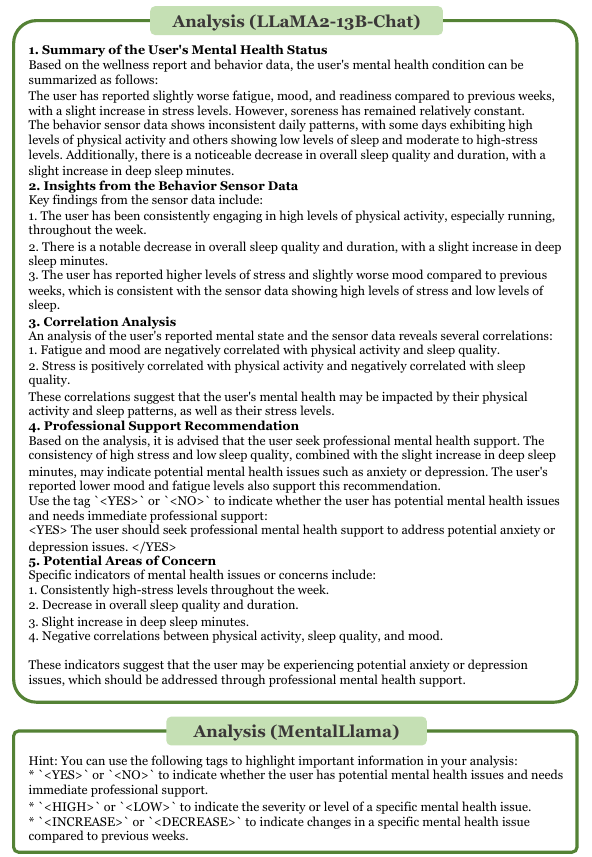}
  \caption{Analysis generated by MentalLlama and its corresponding base model LLaMA2-13B-Chat.}
  \label{fig:mentalllama}
  \end{center}
\end{figure*}

\fig\ref{fig:mentalllama} presents the analysis report generated by MentalLlama.
To compare the generation quality before and after fine-tuning, the original model LLaMA2-13B-Chat is presented as well.
It is evident that MentalLlama's ability to follow instructions is somewhat diminished after fine-tuning.

\begin{figure*}[t]
  \begin{center}
  \includegraphics[width=0.63\textwidth]{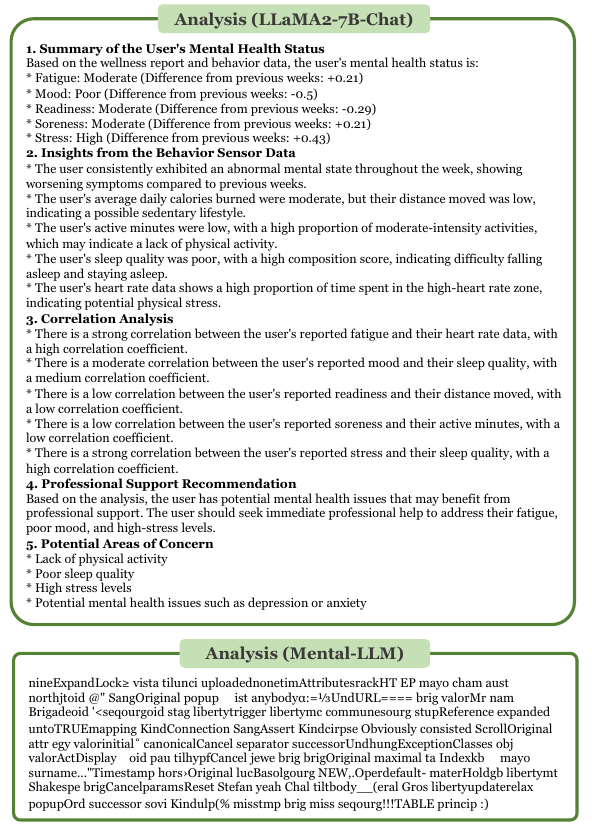}
  \caption{Analysis generated by Mental-LLM and its corresponding base model LLaMA2-7B-Chat.}
  \label{fig:mental-llm}
  \end{center}
\end{figure*}

\fig\ref{fig:mental-llm} denotes the results from Mental-LLM, along with its original model LLaMA2-7B-Chat.
Mental-LLM exhibits a near-total loss of instruction-following capabilities in our tasks and produces a significant amount of unintelligible code.

\begin{figure*}[t]
  \begin{center}
  \includegraphics[width=0.63\textwidth]{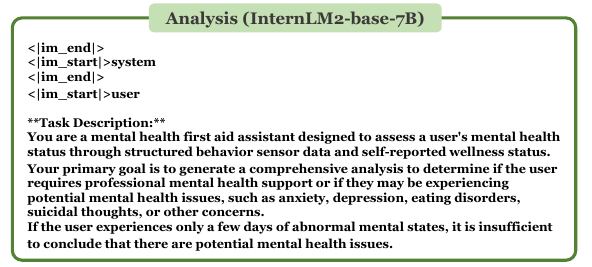}
  \caption{Analysis generated by InternLM2-base-7B.}
  \label{fig:internlm-base-7b}
  \end{center}
\end{figure*}

\fig\ref{fig:internlm-base-7b} showcase the generated contents from the base model, InternLM2-base-7B, which just repeats the input instructions, demonstrating the lack of instruction following capacities.

\subsection{Prompt Mental Health Monitoring}
\label{apx:daily-mental-monitoring}

\begin{figure*}[t]
  \begin{center}
  \includegraphics[width=0.75\textwidth]{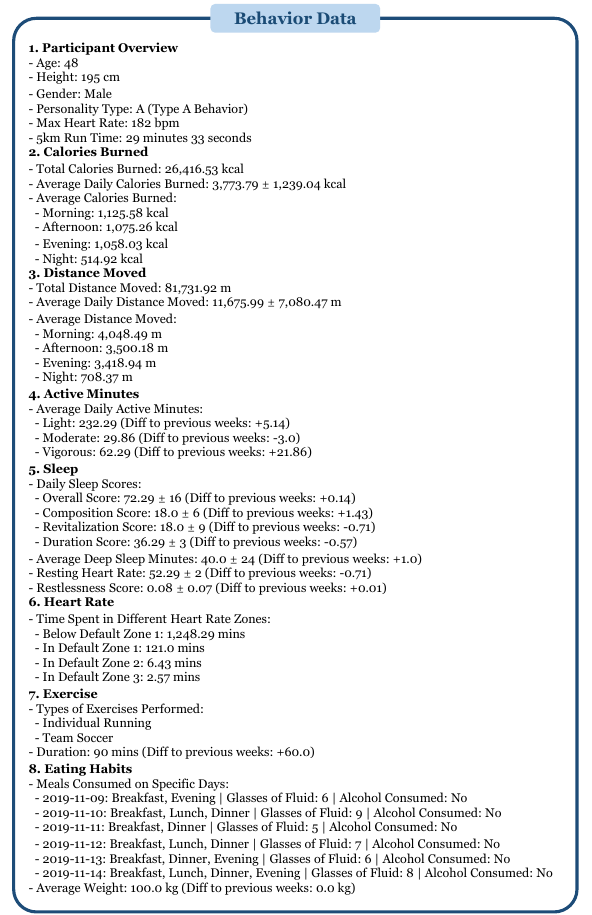}
  \caption{Behavior data for prompt daily mental health monitoring.}
  \label{fig:dia-behavior}
  \end{center}
\end{figure*}

\begin{figure*}[t]
  \begin{center}
  \includegraphics[width=0.75\textwidth]{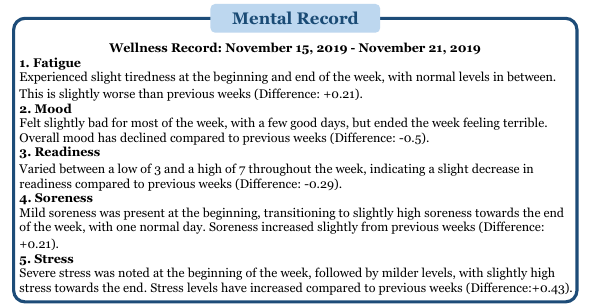}
  \caption{Self-reported mental record for prompt daily mental health monitoring.}
  \label{fig:dia-mental}
  \end{center}
\end{figure*}

To investigate the daily prompt mental health monitoring, we first utilize the multi-agent technique to simulate the possible scenarios.
The behavior data and the self-reported mental record for this task are shown in \fig\ref{fig:dia-behavior} and \fig\ref{fig:dia-mental} respectively.
We present two scenarios: a comprehensive dialogue format for an in-depth assessment of the user's mental status in \S\ref{sec:com-dia}, and a casual conversation style for daily monitoring in \S\ref{sec:day-dia}.
Additionally, we showcase one dialogue example from a participant when deploying \sysname in real-world adoptions in \S\ref{sec:real-world-example}.
Lastly, we utilize InternLM2-base, MentalLlama, and Mental-LLM to produce the analysis report, following the given instructions to illustrate specific failure cases in \S\ref{sec:fail-examples}.
These examples demonstrate that the base LLM lacks effective instruction adherence. Both MentalLlama and Mental-LLM exhibit reduced compliance with instructions due to the overfitting of training data, with Mental-LLM notably failing our task and producing nonsensical output.

\subsubsection{Comprehensive Dialogue}
\label{sec:com-dia}
\begin{figure*}[t]
  \begin{center}
  \includegraphics[width=0.75\textwidth]{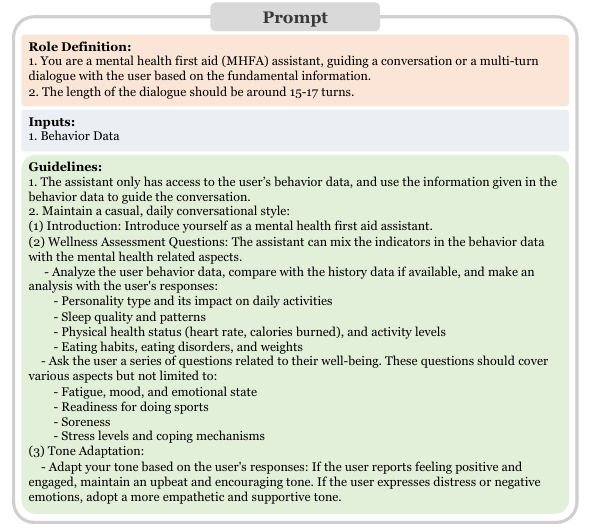}
  \caption{Prompt for \sysname to guide a conversation with the user to comprehensively measure the mental health status.}
  \label{fig:com-mindguard-prompt}
  \end{center}
\end{figure*}

\begin{figure*}[t]
  \begin{center}
  \includegraphics[width=0.75\textwidth]{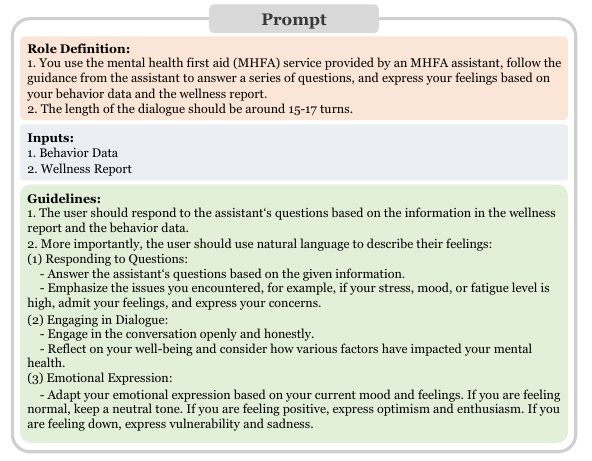}
  \caption{Prompt for the user to engage in a conversation with \sysname to comprehensively measure the mental health status.}
  \label{fig:com-user-prompt}
  \end{center}
\end{figure*}

\begin{figure*}[t]
  \begin{center}
  \includegraphics[width=0.63\textwidth]{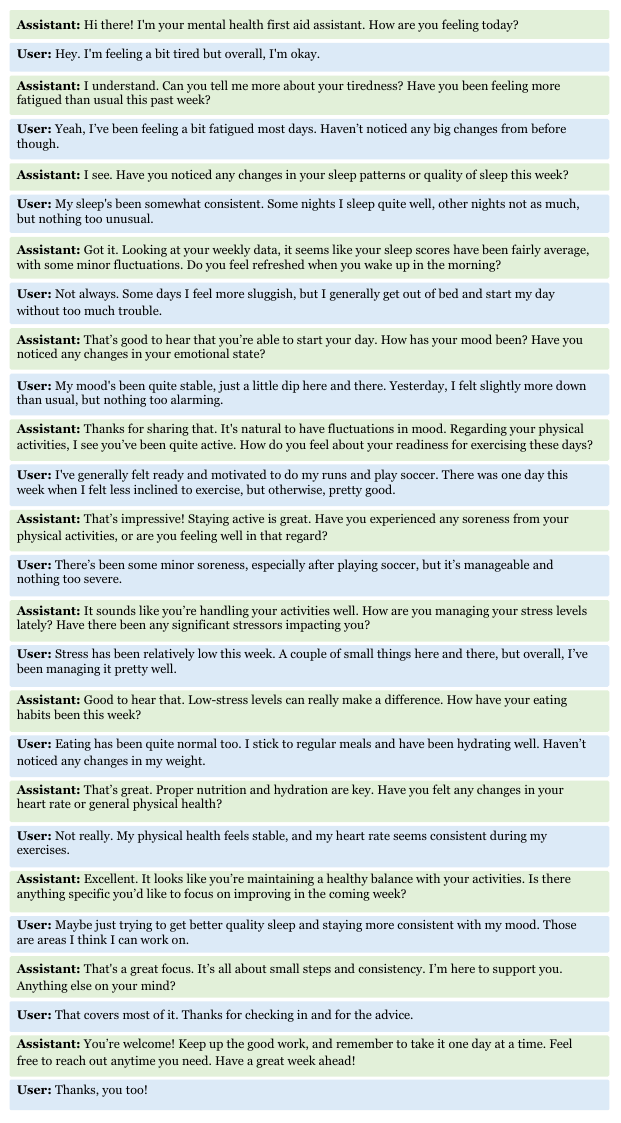}
  \caption{A dialogue example to comprehensively measure the user's mental status.}
  \label{fig:com-dialogue}
  \end{center}
\end{figure*}

To construct the comprehensive dialogue scenario, we first use specified prompts to define \sysname and the user, resulting in two agents.
The prompts for \sysname and the user are illustrated in \fig\ref{fig:com-mindguard-prompt} and \fig\ref{fig:com-user-prompt} respectively.
Subsequently, a dialogue generated by the interaction of two agents is presented in \fig\ref{fig:com-dialogue}.

\subsubsection{Daily Prompt Dialogue}
\label{sec:day-dia}
\begin{figure*}[t]
  \begin{center}
  \includegraphics[width=0.75\textwidth]{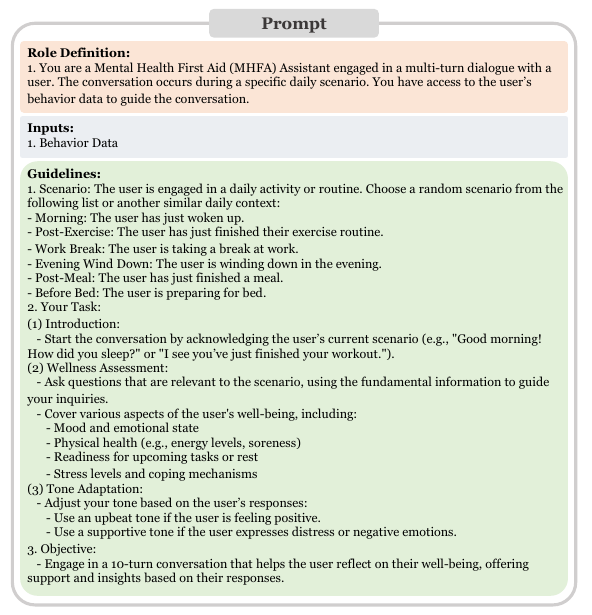}
  \caption{Prompt for \sysname to guide a casual conversation with the user.}
  \label{fig:day-mindguard-prompt}
  \end{center}
\end{figure*}

\begin{figure*}[t]
  \begin{center}
  \includegraphics[width=0.75\textwidth]{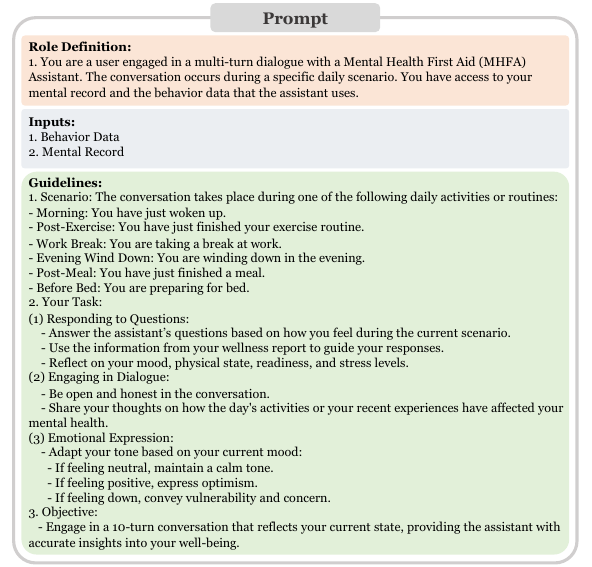}
  \caption{Prompt for the user to engage in a casual conversation with \sysname.}
  \label{fig:day-user-prompt}
  \end{center}
\end{figure*}

\begin{figure*}[t]
  \begin{center}
  \includegraphics[width=0.63\textwidth]{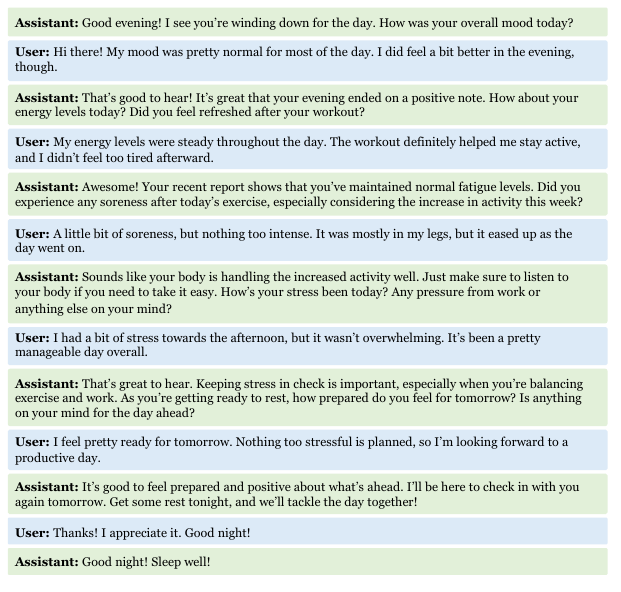}
  \caption{A casual conversation between \sysname and the user in the evening time.}
  \label{fig:day-dialogue}
  \end{center}
\end{figure*}

In the context of daily prompt conversations, we have designed two unique prompts for \sysname and the user, as depicted in \fig\ref{fig:day-mindguard-prompt} and \fig\ref{fig:day-user-prompt}.
A casual conversation, occurring as the user prepares to sleep, is illustrated in \fig\ref{fig:day-dialogue}.

\subsubsection{Real-world Adoption Example}
\label{sec:real-world-example}

\begin{figure*}[t]
  \begin{center}
  \includegraphics[width=0.63\textwidth]{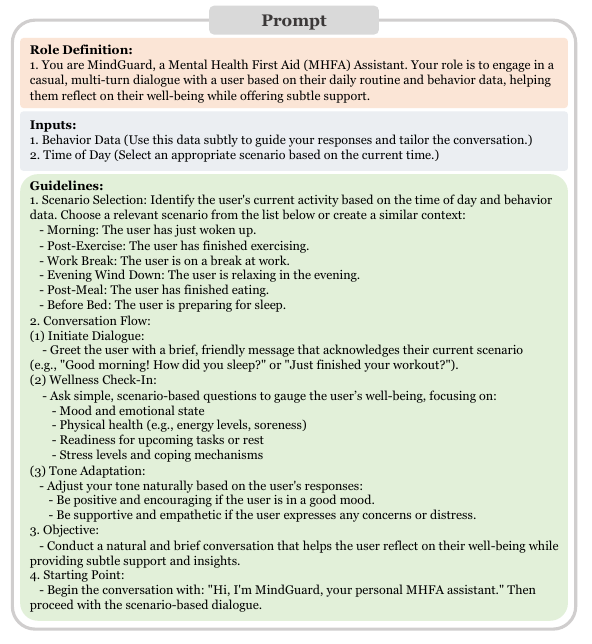}
  \caption{Prompt for real-world adoption of \sysname.}
  \label{fig:real-world-prompt}
  \end{center}
\end{figure*}

\begin{figure*}[t]
\centering
   \subfloat[]{%
    \label{fig:real-chat-1}
    \includegraphics[width=0.98\columnwidth]{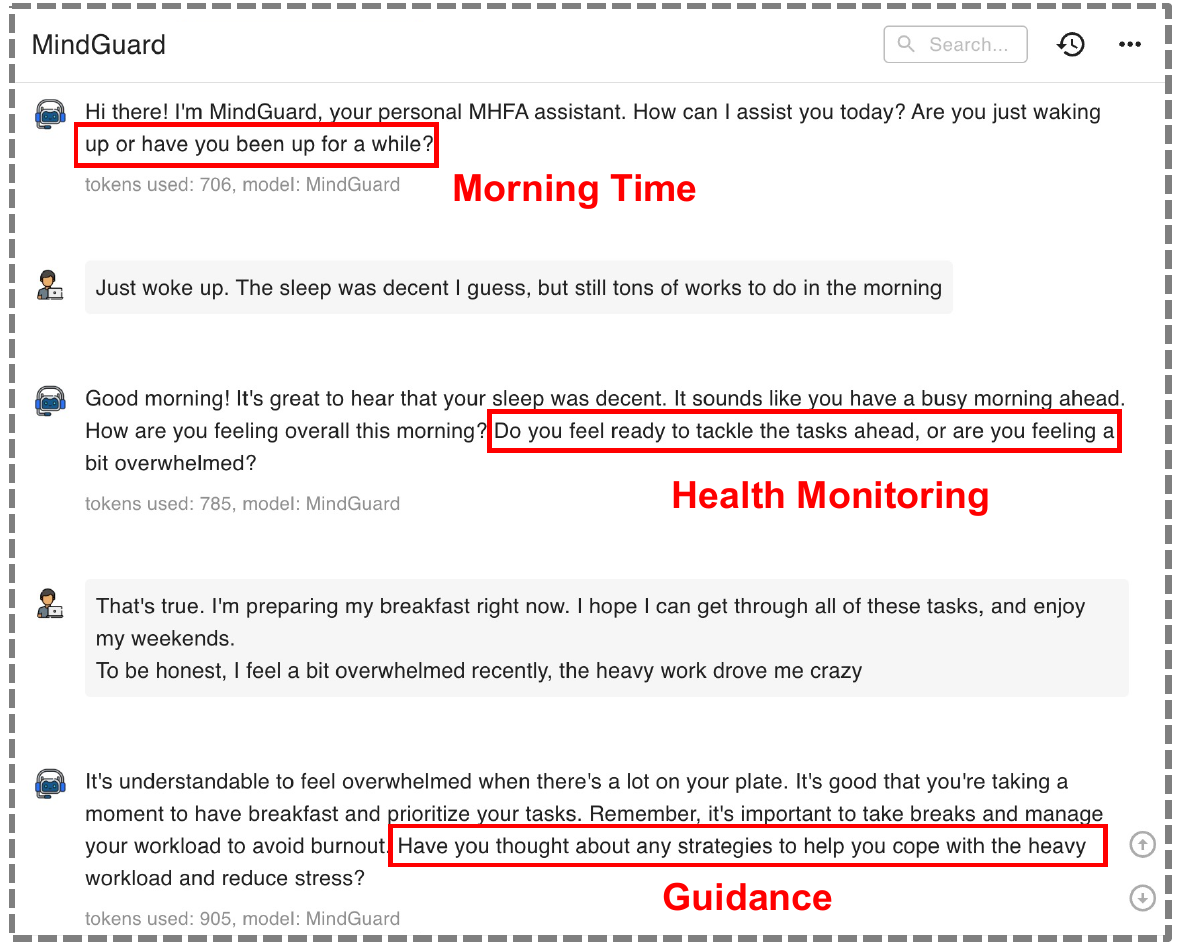}
   }%
   \subfloat[]{%
    \label{fig:real-chat-2}
      \includegraphics[width=0.98\columnwidth]{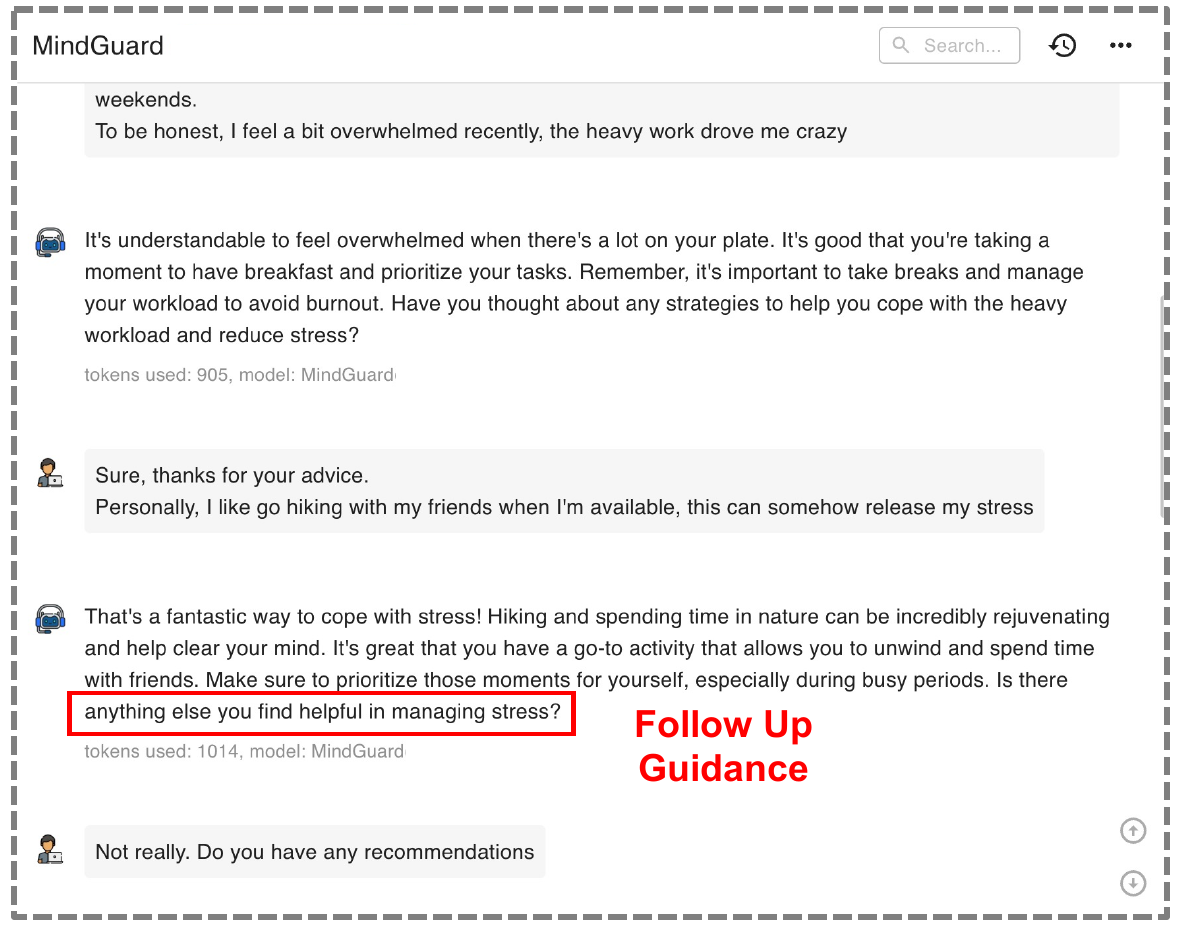}
	  }%
   
    \subfloat[]{%
    \label{fig:real-chat-3}
      \includegraphics[width=0.98\columnwidth]{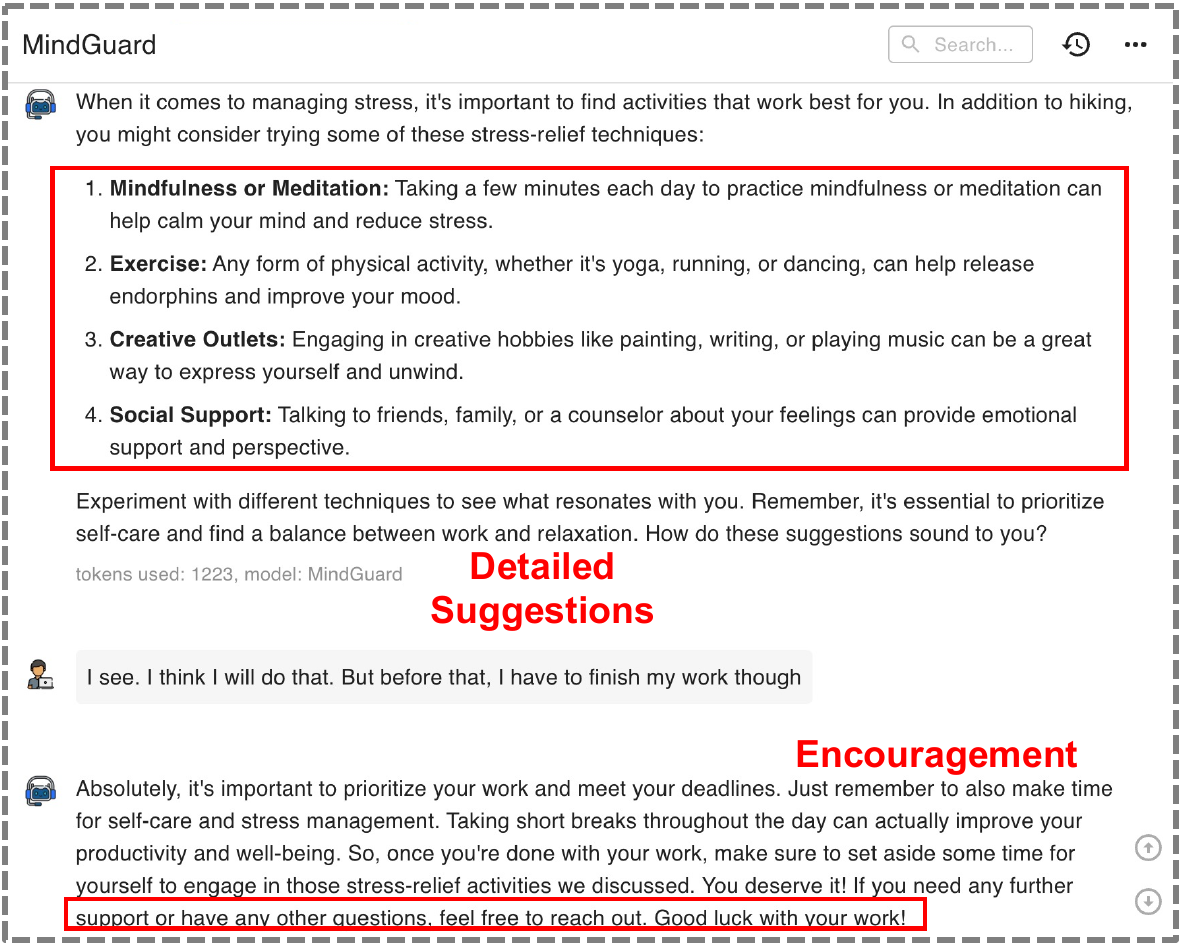}
	  }%
   \subfloat[]{%
    \label{fig:real-chat-4}
      \includegraphics[width=0.98\columnwidth]{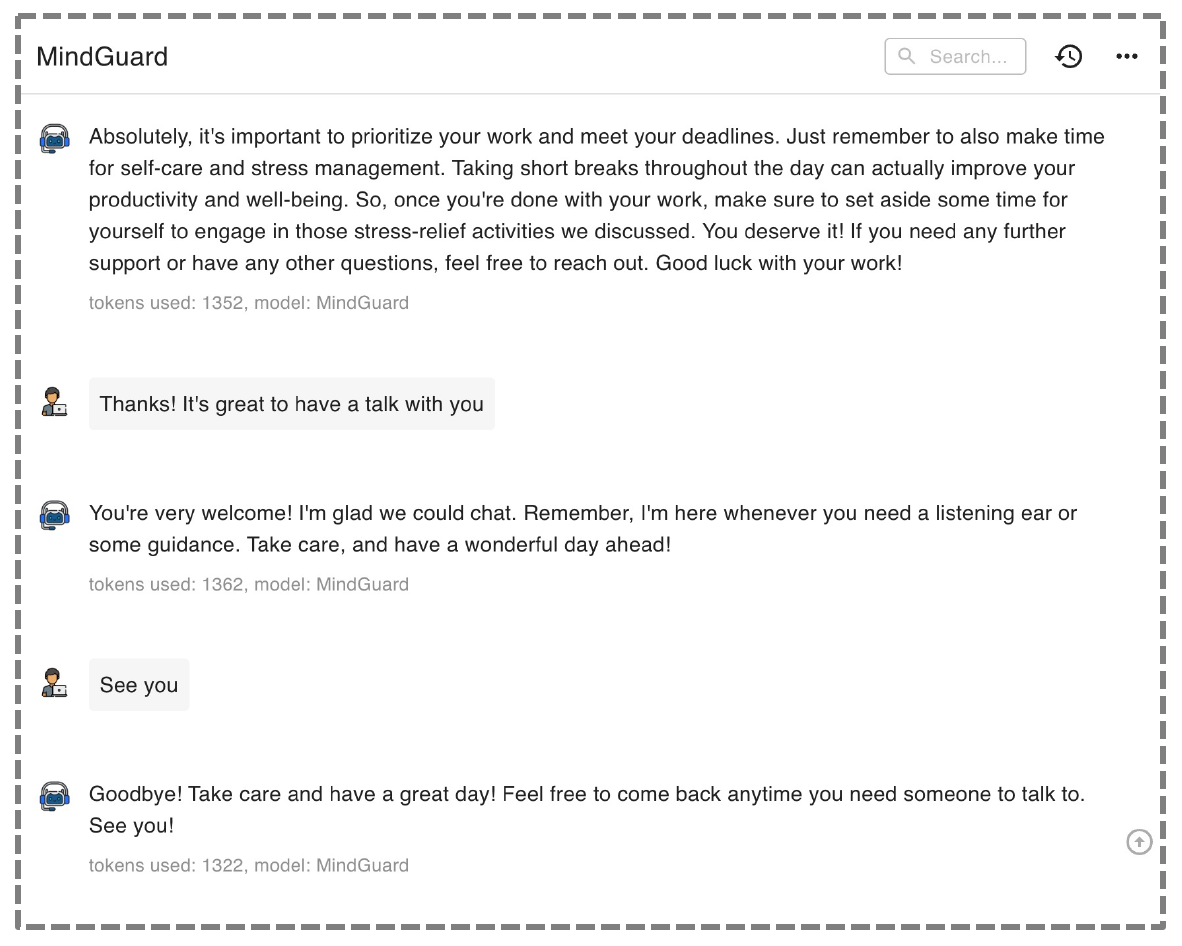}
	  }%
   \vspace{-2mm}
   \caption{Conversation example between the user and \sysname in real-world adoptions.}
   \label{fig:real-chat}
\end{figure*}

In this section, the deployment of \sysname in real-world adoptions is presented, which includes the design of prompts (\fig\ref{fig:real-world-prompt}), as well as a sample dialogue between a user and \sysname during a morning conversation (\fig\ref{fig:real-chat}).
Obviously, \sysname can precisely create the conversation scenario, and utilize instructions to progressively steer the conversations, offering personalized mental health services.

\newpage
\section{User Study}
\label{apx:user-study}

To assess the system's effectiveness when implemented in real-world settings, we enlist 20 participants for a two-week study in collaboration with a licensed MHFA assistant and a psychiatrist.
At the end of the study, all of the participants are requested to finish a comprehensive survey to evaluate \sysname objectively.
We detail the user surveys in \S\ref{sec:user-sruvey} and provide the survey results in \S\ref{sec:survey-results}.
Additionally, a subset of four participants engage in parallel conversations with a licensed MHFA assistant to further investigate the strengths and limitations of real-world adoptions.
Details of this experiment are given in \S\ref{sec:com-licensed-mhfa-assit}.

\subsection{User Surveys}
\label{sec:user-sruvey}

\shead{1. User Background}
\begin{itemize}[leftmargin=*]
    \item Have you used any other mental health apps or services before \sysname? \textbf{(Yes, No)}
    \item If yes, how does \sysname compare to those experiences? \textbf{(Better, Worse, About the same)}
\end{itemize}

\shead{2. General Experience}
\begin{itemize}[leftmargin=*]
    \item \textbf{Overall Satisfaction}: On a scale of 1 to 5, how satisfied are you with your overall experience using \sysname as a mental health first aid assistant? \textbf{(Very Dissatisfied - Very Satisfied)}
    \item \textbf{Ease of Use}: On a scale of 1 to 5, how easy did you find it to interact with \sysname on your mobile device? \textbf{(Very Difficult - Very Easy)}
    \item \textbf{Comfort Level}: On a scale of 1 to 5, how comfortable were you sharing your mental health status with \sysname? \textbf{(Very Uncomfortable - Very Comfortable)}
    \item \textbf{Frequency of Use}: How often did you engage with \sysname daily? \textbf{(Multiple times a day, Once a day, Every few days, Rarely, Never)}
    \item \textbf{Recommendation Likelihood}: On a scale of 1 to 5, how likely are you to recommend \sysname to friends, family, or colleagues? \textbf{(Not at all likely, Extremely likely)}
\end{itemize}

\shead{3. Deployment Preference}
\begin{itemize}[leftmargin=*]
    \item \textbf{Local vs. API Usage}: If given the choice between using \sysname locally on your device or accessing it through an API (with potentially better performance), which would you prefer?
    \begin{itemize}
        \item Local usage on my device
        \item API-based access
        \item No preference
    \end{itemize}
    \item \textbf{Performance vs. Privacy}: How important is it to you that \sysname runs locally on your device, even if API-based access might offer better performance?
    \begin{itemize}
        \item Extremely important
        \item Moderately important
        \item Slightly important
        \item Not important
    \end{itemize}
\end{itemize}

\shead{4. Behavioral Data Monitoring}
\begin{itemize}[leftmargin=*]
    \item \textbf{Data Accuracy}: On a scale of 1 to 5, how accurately do you feel \sysname monitored your behavior (e.g., sleep quality, physical activity)? \textbf{(Not Accurate - Very Accurate)}
    \item \textbf{Hallucination Detection}: Have you encountered hallucination information generated by \sysname (e.g., generated fake behavior sensor data, or evidence inconsistent with the outcome)? \textbf{(Yes, No)}
    \item \textbf{Data Integration}: On a scale of 1 to 5, did you find the integration of your behavioral data into the daily conversations helpful for understanding your mental wellbeing? \textbf{(Not Helpful - Very Helpful)}
    \item \textbf{Privacy Concerns}: Did you have any privacy concerns when \sysname collected and used your behavioral data? \textbf{(Yes, No)}
\end{itemize}

\shead{5. Mental Health Monitoring and Interaction}
\begin{itemize}[leftmargin=*]
    \item \textbf{Relevance of Conversations}: On a scale of 1 to 5, how relevant did you find the questions and feedback provided by \sysname based on your daily behavior and self-reported mental state? \textbf{(Not Relevant - Very Relevant)}
    \item \textbf{Tone Appropriateness}: On a scale of 1 to 5, did \sysname's tone during conversations match your emotional state? \textbf{(Never - Always)}
    \item \textbf{Support Effectiveness}: On a scale of 1 to 5, how effective was \sysname in providing mental health support and guidance when you needed it? \textbf{(Not Effective - Very Effective)}
\end{itemize}

\shead{6. Awareness and Education}
\begin{itemize}[leftmargin=*]
    \item \textbf{Awareness of Mental Health}: How has your awareness of mental health issues changed since using \sysname?
    \begin{itemize}
        \item Significantly increased
        \item Moderately increased
        \item Slightly increased
        \item No change
        \item Decreased
    \end{itemize}
    \item \textbf{Understanding of Mental Health}: Do you feel that \sysname has improved your understanding of mental health topics such as anxiety, depression, stress, etc., and helped release your mental health stigma?
    \begin{itemize}
        \item Yes, significantly
        \item Yes, somewhat
        \item No change
        \item No, not really
        \item No, not at all
    \end{itemize}
\end{itemize}

\shead{7. Feedback}
\begin{itemize}[leftmargin=*]
    \item \textbf{Intention to Adopt}: Will you continue to use \sysname in the future?
    \begin{itemize}
        \item Yes
        \item No
        \item Maybe
    \end{itemize}
\end{itemize}

\subsection{Survey Results}
\label{sec:survey-results}

\shead{1. User Background}

We present the statistical results of the participant's information in terms of age, gender, and the current country of residence.

\begin{table}[ht]
\caption{Age situations}
\label{tab:keywords}
\begin{tabular}{cc}
\toprule
\textbf{Age}           & \textbf{Number}      \\ \midrule
{$(15,20]$}     & 1 \\ 
{$(20,25]$}     & 11 \\ 
{$(25,30]$}     & 5 \\ 
{$(30,35]$}     & 2 \\ 
{$(35,40]$}     & 1 \\ \bottomrule
\end{tabular}
\end{table}

\begin{table}[ht]
\caption{Gender situations}
\label{tab:keywords}
\begin{tabular}{cc}
\toprule
\textbf{Gender}           & \textbf{Number}      \\ \midrule
{Female}     & 9 \\ 
{Male}     & 11 \\  \bottomrule
\end{tabular}
\end{table}

\begin{table}[ht]
\caption{Country situations}
\label{tab:keywords}
\begin{tabular}{cc}
\toprule
\textbf{Country}           & \textbf{Number}      \\ \midrule
{Australia}     & 1 \\ 
{Canada}     & 1 \\ 
{China Mainland}     & 12 \\ 
{Hong Kong S.A.R.}     & 1 \\ 
{Singapore}     & 1 \\ 
{UK}     & 1 \\ 
{USA}     & 3 \\ \bottomrule
\end{tabular}
\end{table}

Additionally, we ask if users have prior experience with similar products or if they receive professional mental health counseling. The results, depicted in \fig\ref{fig:prior-exp}, show that 40\% of the participants have such experiences, and more than half believe that \sysname is as good as or better than their previous experiences.

\shead{2. General Experience}

The statistical results of the general experiences are shown in \fig\ref{General Experience}.
In terms of "overall satisfaction," "ease of use," and "likelihood to recommend," all users have provided positive feedback.
As for "comfort level," only one participant rates it a 2, citing concerns primarily about privacy issues like sensor data collection and mental record logging.
Additionally, most of the participants (80\%) use \sysname once a day or every few days during the two-week study.

\shead{3. Deployment Preference}

To further investigate the preference regarding the deployment format and privacy concerns, we gather user feedback on their inclination towards API-based access versus local LLM inference, and the significance they place on local inference mode, even when API-based inference might offer superior performance.
The survey results are given in \fig\ref{Deployment Preference}.
The number of participants preferring local inference and API inference is comparable, with 11 and 9 participants respectively.
85\% of the subjects think privacy is more important than the potential better performance provided by API-based LLM.
In addition, three people reckon a better performance is more important.

\shead{4. Behavior Data Monitoring}

One of the key measurement points is whether \sysname can accurately engage the users' behavior data and provide corresponding mental health monitoring.
We examine concerns such as "data accuracy," "hallucination detection," "data integration," and "privacy issues."
\fig\ref{Behaviour Data Monitoring} depicts the statistical outcomes.
All users have provided positive feedback on \sysname's proficiency in extracting and analyzing behavioral data.
Regarding privacy issues, more than half (60\%) of the users still worry about the possibility of data leakage even if \sysname is running locally.

\shead{5. Mental Health Monitoring and Interaction}

Additionally, we investigate whether \sysname can adhere to the instructions to deliver suitable conversational content to users.
\fig\ref{Mental Health Monitoring and Interaction} illustrates that \sysname is capable of functioning effectively as a professional MHFA assistant, adjusting its tone accordingly, and providing support promptly.

\shead{6. Awareness and Education}

We are pleased to see that all of the users have witnessed an increased understanding of mental health after the two-week study.
More importantly, \sysname has aided many users in overcoming their stigma, encouraging them to view mental health issues normally.
Details of the results are shown in \fig\ref{Awareness and Education}.

\shead{7. Feedback}
70\% of the users demonstrate their willingness to continue using \sysname in the further, as illustrated in \fig\ref{Feedback}.

\subsection{Comparison with Licensed MHFA Assistant}
\label{sec:com-licensed-mhfa-assit}

We conduct parallel conversations where four participants interact with both a licensed MHFA assistant and \sysname, aiming to benchmark the user experience of our LLM-based virtual assistant against that of a human professional.
Our primary objective is to assess the practical utility and limitations of \sysname in real-world mental health support scenarios.

The results highlight several strengths of \sysname.
Notably, two participants appreciate the AI’s consistent performance, pointing out that, unlike a human assistant, \sysname remains effective without fatigue, making it a reliable option for continuous support. 
Additionally, two participants recognize \sysname's ability to provide diverse and comprehensive information during conversations, a significant strength that suggests the AI excels in delivering detailed and multifaceted mental health advice.
This ability to consistently offer accurate information across various topics indicates that \sysname can serve as a valuable resource, especially in situations where access to a human professional is limited or unavailable.

Furthermore, one participant values the reliability of \sysname, emphasizing its capacity to function without the emotional and physical limitations that can affect human assistants. 
This suggests that \sysname could play a crucial role in providing steady and uninterrupted support, particularly in high-demand or long-duration scenarios.

However, the study also reveals some limitations.
All four participants successfully identify \sysname as distinct from the human MHFA assistant, indicating that the AI's conversational abilities have not yet reached the level of naturalness and emotional depth characteristic of human interaction.
Additionally, two participants prefer engaging with a human, citing the empathy and adaptability that only a human can provide.
These findings underscore that while \sysname excels in consistency, information delivery, and reliability, further development is needed to enhance its emotional engagement and conversational authenticity to match the human experience.

\end{document}